\documentclass[nohyperref]{article}

\usepackage{microtype}
\usepackage{graphicx}
\usepackage{booktabs} 

\usepackage{hyperref}
\usepackage[acronym]{glossaries}
\usepackage{array, multirow}
\usepackage{subfloat}
\usepackage{subcaption}
\usepackage{tikz}
\usetikzlibrary{shapes,snakes}
\usepackage{enumitem}
\usepackage[group-separator={,},group-minimum-digits=4]{siunitx}
\usepackage{xcolor}
\usepackage{wrapfig}
\usepackage{xfrac}
\usepackage{tabularx}
\usepackage{url}

\usepackage{amsmath,amsfonts,bm}

\def\eqref#1{equation~\ref{#1}}

\def\1{\bm{1}}

\def\vh{{\bm{h}}}

\def\vs{{\bm{s}}}

\def\vx{{\bm{x}}}

\def\vz{{\bm{z}}}

\def\mW{{\bm{W}}}

\DeclareMathAlphabet{\mathsfit}{\encodingdefault}{\sfdefault}{m}{sl}
\SetMathAlphabet{\mathsfit}{bold}{\encodingdefault}{\sfdefault}{bx}{n}

\usepackage{booktabs}

\usepackage[accepted]{icml2022}

\usepackage{amsmath}
\usepackage{amssymb}
\usepackage{mathtools}
\usepackage{amsthm}

\usepackage[capitalize,noabbrev]{cleveref}

\theoremstyle{plain}

\theoremstyle{definition}

\theoremstyle{remark}

\usepackage[textsize=tiny]{todonotes}

\newacronym{dnn}{DNN}{Deep Neural Network}
\newacronym{cnn}{CNN}{Convolutional Neural Network}
\newacronym{nn}{NN}{neural network}
\newacronym{ml}{ML}{Machine Learning}
\newacronym{vtab}{VTAB}{Visual Task Adaptation Benchmark}
\newacronym{tl}{TL}{transfer learning}
\newacronym{ft}{\textsc{FineTuning}}{fine-tuning}
\newacronym{lp}{\textsc{Linear}}{linear probing}
\newacronym{ood}{OOD}{out-of-distribution}
\newacronym{h2t}{\textsc{Head2Toe}}{head-to-toe probing} 
\newacronym{batchnorm}{BatchNorm}{Batch Normalization}

\newcommand\lp{\textsc{Linear}}
\newcommand\ft{\textsc{FineTuning}}
\newcommand\scr{\textsc{Scratch}}
\newcommand\imgnet{ImageNet-2012}

\definecolor{natural}{HTML}{648FFF}
\definecolor{specialized}{HTML}{DC267F}
\definecolor{structured}{HTML}{362682}
\definecolor{all}{HTML}{FE6100}

\icmltitlerunning{Head2Toe: Utilizing Intermediate Representations for Better Transfer Learning}

\begin{document}

\twocolumn[
\icmltitle{Head2Toe: Utilizing Intermediate Representations \\for Better Transfer Learning}




\begin{icmlauthorlist}
\icmlauthor{Utku Evci}{goog}
\icmlauthor{Vincent Dumoulin}{goog}
\icmlauthor{Hugo Larochelle}{goog}
\icmlauthor{Michael C. Mozer }{goog}
\end{icmlauthorlist}

\icmlaffiliation{goog}{Google Research, Brain Team}

\icmlcorrespondingauthor{Utku Evci}{evcu@google.com}

\icmlkeywords{Machine Learning, ICML}

\vskip 0.3in
]



\printAffiliationsAndNotice{} 

\begin{abstract}
Transfer-learning methods aim to improve performance in a data-scarce target  domain using a model pretrained on a data-rich source domain. A cost-efficient strategy, \emph{linear probing}, involves freezing the source model and training a new classification head for the target domain. This strategy is outperformed by a more costly but state-of-the-art method---\emph{fine-tuning} all parameters of the source  model to  the target domain---possibly because fine-tuning allows the model to leverage  useful information from intermediate layers which is otherwise discarded by the previously trained later layers. We explore the hypothesis that these intermediate layers  might be directly exploited. We propose a method, \emph{Head-to-Toe probing} (\textsc{Head2Toe}), that selects features from all layers of the source model to train a classification head for the target domain. In evaluations on the \gls{vtab}, Head2Toe matches performance obtained with fine-tuning on average while reducing training and storage cost a hundred fold or more, but critically, for out-of-distribution transfer, Head2Toe outperforms fine-tuning\footnote{We open source our code at \url{https://github.com/google-research/head2toe}}.
\end{abstract}

\section{Introduction}
\label{introduction}
Transfer learning is a widely used method for obtaining strong performance in a variety of tasks where training data is scarce \citep[e.g.,][]{zhu2020transfer,alyafeai2020survey,zhuang2020comprehensive}. 
A well-known recipe for transfer learning involves the supervised or unsupervised pretraining of a model on a \emph{source} task with a large training dataset (also referred to as \emph{upstream training}). After pretraining, the model's output head is discarded, and the rest of the network is used to obtain a feature embedding, i.e., the output of what was formerly the penultimate layer of the network. When transferring to a \emph{target} task, a new output head is trained on top of the feature extractor (\emph{downstream training}). This approach makes intuitive sense: if a linear combination of embedding features performs well on the source task and the source and target domains are similar, we would expect a different linear combination of features to generalize to the target domain.

\begin{figure}[t]
\centering
\includegraphics[width=0.7\columnwidth]{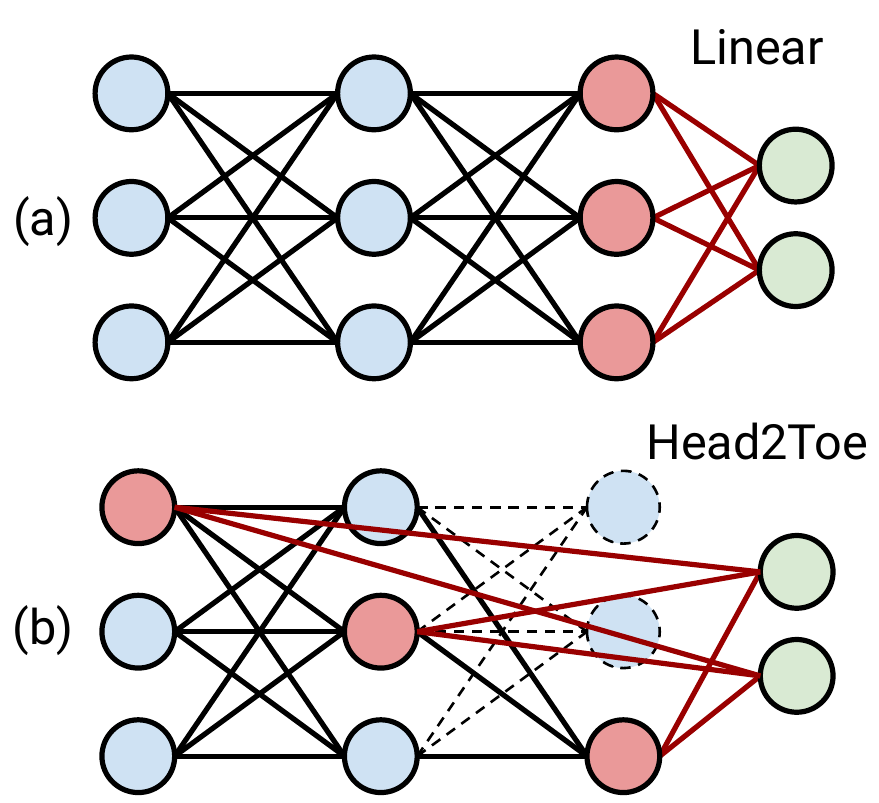}
\caption{\textbf{(a)} Whereas \gls{lp} utilizes only the last layer for transfer learning, \textbf{(b)} \gls{h2t} selects the most useful features from the entire network and trains a linear head on top. }
\label{fig:intro}
\vspace{-.7em}
\end{figure}

This approach of training a new output head, which we refer to as \emph{\gls{lp}}, often yields
significant improvements in performance on the target task over training the network from
scratch \citep{Kornblith2019DoBI}. An alternative to \gls{lp} is \emph{\gls{ft}}, which uses target-domain data to adapt all weights in the feature extractor together with the new output head. This procedure requires running forward and backward passes through the entire network at each training step and therefore its per-step cost is significantly higher than \gls{lp}. Furthermore, since the entire network is fine-tuned, the entire set of new weights needs to be stored for every target task, making \gls{ft} impractical when working on edge devices or with a large number of target tasks. However, \gls{ft} is often preferred over \gls{lp} since it consistently leads to better performance on a variety of target tasks even when data is scarce~\citep{Zhai2019TheVT}.

\gls{ft}'s superior generalization in the low-data regime is counterintuitive
given that the number of model parameters to be adapted is often large relative to
the amount of available training data.  How does \gls{ft} learn from few examples successfully? We conjecture that \gls{ft} better leverages existing
internal representations rather than discovering entirely new representations;
\gls{ft} exposes existing features buried deep in the net for use by the classifier.
Under this hypothesis, \emph{features needed for transfer are already present in the pretrained network and might be identified directly without fine-tuning the backbone itself}. In \cref{subsec:taylor}, we argue that \gls{ft} can be approximated by a linear probe operating on the intermediate features of a network, thus enabling state-of-the-art transfer performance with significantly less cost.

\begin{figure}[t]
\centering
\includegraphics[width=\columnwidth]{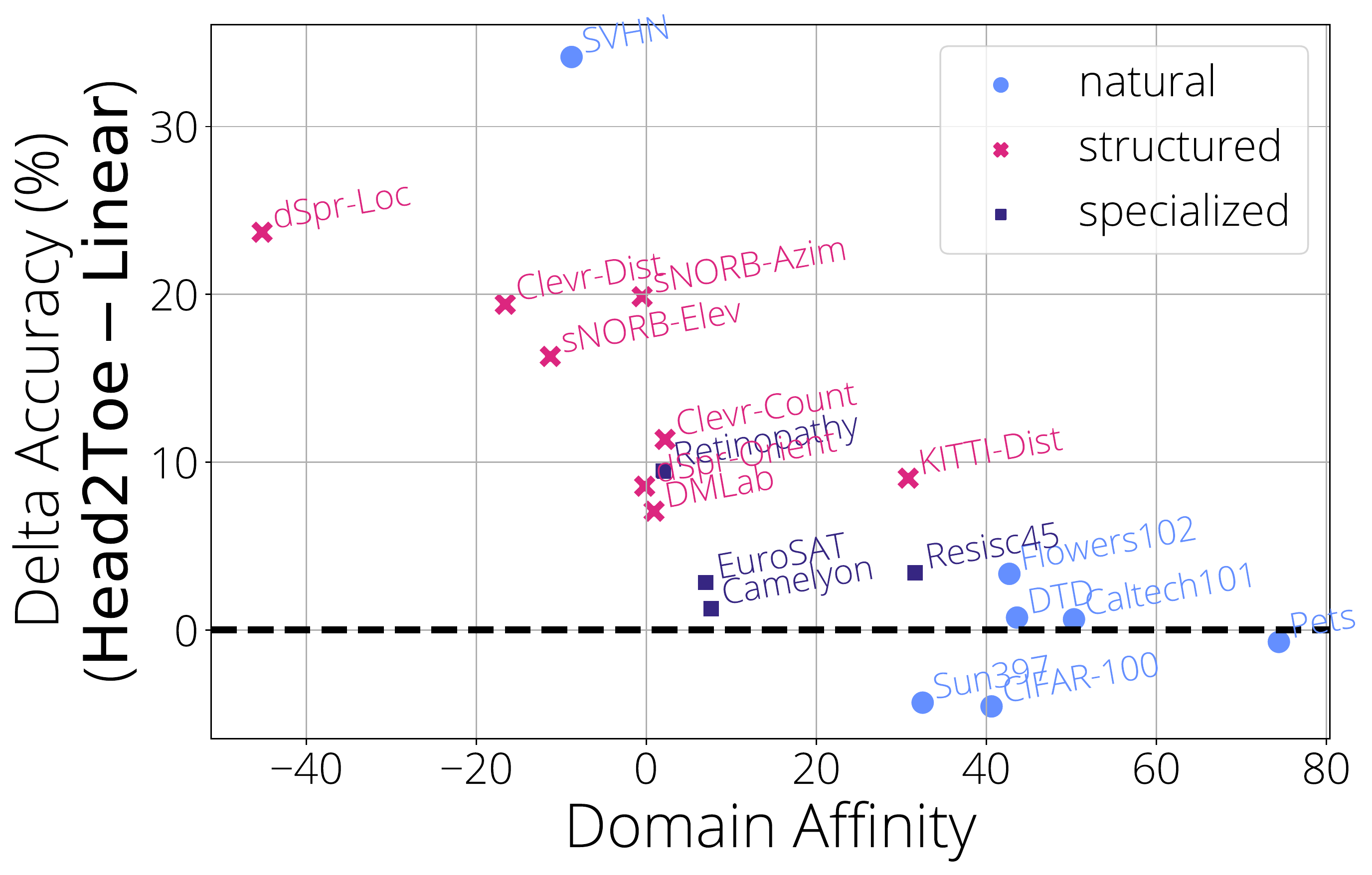}
\caption{Improvement in accuracy due to \gls{h2t} over \gls{lp} for the \gls{vtab} target domains,
arranged on the abscissa by domain affinity (\cref{sec:prelim}),
proxy for OOD-ness of the target relative to the source domain (ImageNet). The more OOD domains (left side of the plot) benefit most from \gls{h2t}.}
\label{fig:intro2}
\end{figure}

In this work, we propose and explore methods for selecting useful features from \emph{all} layers of a pretrained net, including the embedding, and then applying the \gls{lp} transfer approach to the constructed representation. We compare the standard approach (\cref{fig:intro}a) to our approach, called \gls{h2t} (\cref{fig:intro}b). \gls{h2t} shows significant improvements over \lp{} (\cref{fig:intro2}) and matches \gls{ft} performance on average. Our key contributions are as follows:

\vspace{-.1in}
\begin{enumerate}
    \item We observe a strong correlation between the degree to which a target domain is \gls{ood} with respect to the source domain and the benefit of incorporating intermediate representations in \gls{lp} (\cref{fig:intro2} and \cref{sec:h2t:2}), corroborating observations made in \citet{Adler2020CrossDomainFL}.
    \item We introduce \gls{h2t}, an efficient transfer learning method for selecting relevant features from intermediate representations (\cref{sec:h2t:3}).
    \item On the \gls{vtab} collection of data sets, we show that \gls{h2t} outperforms \gls{lp} and matches the performance of the more 
    computationally costly \gls{ft} with only 0.6\% of the training FLOPs and 1\% of the storage cost (\cref{sec:exps}).  
    \item Critically, \gls{h2t} outperforms \gls{ft} on \gls{ood} target domains. If a practitioner can make an educated guess about whether a target domain is \gls{ood} with respect to a source, using \gls{h2t} improves on the state-of-the-art for transfer learning.
\end{enumerate}

\vspace{-.15in}
\section{Preliminaries}
\label{sec:prelim}
\begin{figure*}[ht]
\centering
\begin{minipage}{.09\textwidth}
\includegraphics[width=\columnwidth]{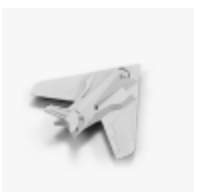}
\includegraphics[width=\columnwidth]{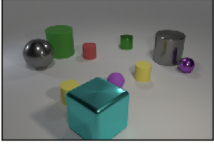}
\includegraphics[width=\columnwidth]{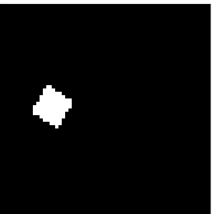}
\end{minipage}%
\begin{minipage}{.81\textwidth}
\centering
\includegraphics[width=0.9\columnwidth]{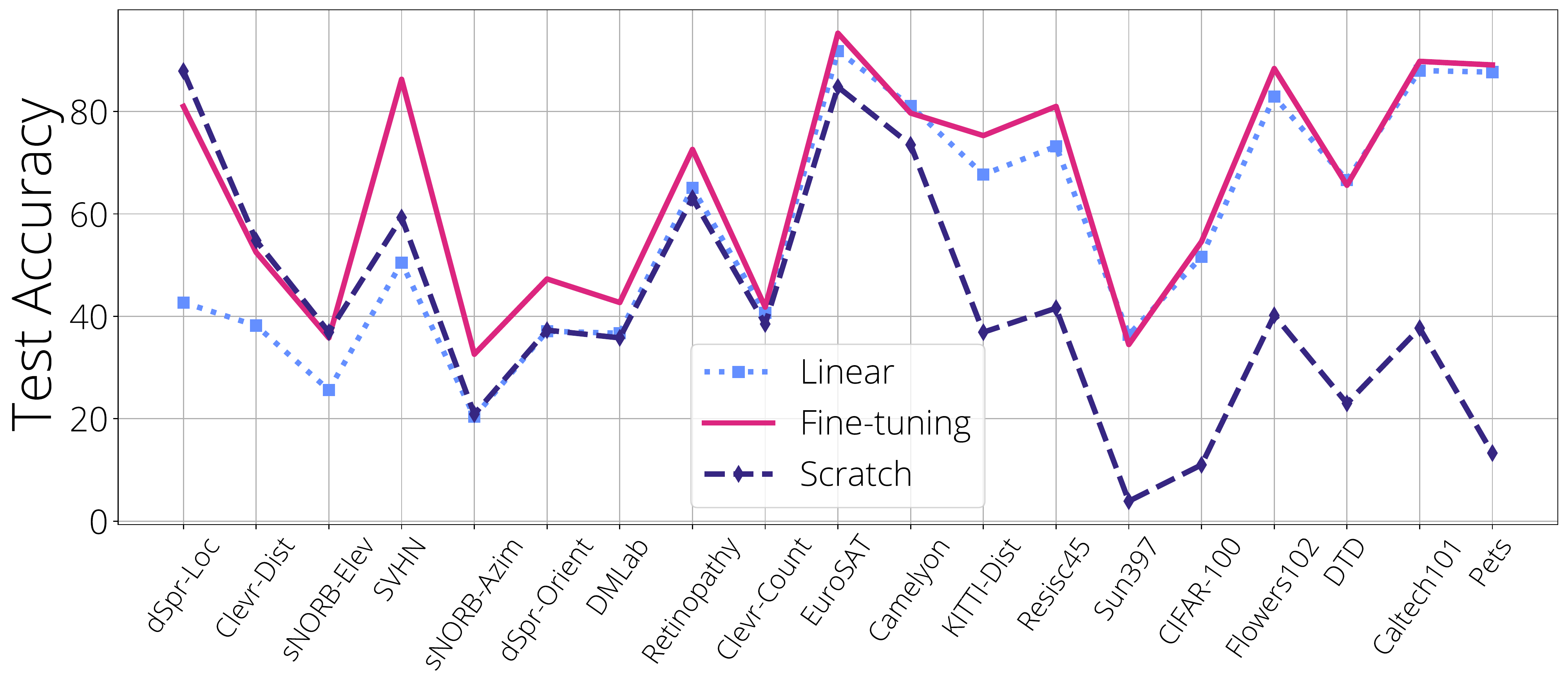}
\end{minipage}%
\begin{minipage}{.09\textwidth}
\includegraphics[width=\columnwidth]{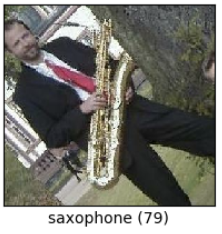}
\includegraphics[width=\columnwidth]{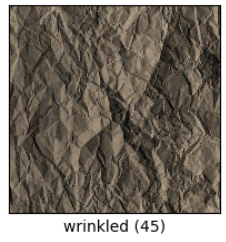}
\includegraphics[width=\columnwidth]{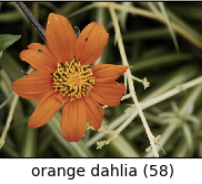}
\end{minipage}%
\caption{\textbf{Characterizing out-of-distribution (far) domains} Generalization performance of various baselines on the VTAB-1k benchmark using ResNet-50 architecture and 240 image size. The architecture is pretrained on \imgnet{} for the \gls{tl} baselines. Datasets (and the three groups of the benchmark) are ordered according to their Domain Affinity scores in ascending order from left to right. Examples from the left- and right-most datasets are also shown on corresponding sides.}
\label{fig:ood_metric}
\end{figure*}

\paragraph{Source domain and backbone models.} In our experiments, we use source models pretrained on \imgnet~\citep{imagenet}, a large scale image classification benchmark with 1000 classes and over 1M natural images. We benchmark \gls{h2t} using convolutional  \citep[ResNet-50, ][]{resnet} and attention-based \citep[ViT-B/16, ][]{vit} architectures pretrained on \imgnet.  
\vspace{-0.5em}
\paragraph{Target domains.} In this work, we focus on target tasks with few examples (i.e., \textit{few-shot}) and use \acrlong{vtab}-1k \citep{Zhai2019TheVT} to evaluate different methods. \acrshort{vtab}-1k consists of 19 different  classification tasks, each having between 2 to 397 classes and a total of 1000 training examples. The domains are grouped into three primary categories: (1) natural images  (\textit{natural}), (2) specialized images using non-standard cameras (\textit{specialized}), and (3) rendered artificial images (\textit{structured}). 
\vspace{-.5em}

\paragraph{Characterizing out-of-distribution (far) domains.} 
\citet{Adler2020CrossDomainFL} use the difference in Fréchet inception distance (FID) between two domains to characterize how far OOD domains are. However, we need a metric which reflects not only changes in $p(x)$---as FID does---but also changes in the task $p(y \mid x)$ itself. Consider the relationship between source and target domains. If the domain overlap is high, then features extracted for linear classification in the source domain should also be relevant for linear classification in the target domain, and \lp{} should yield performance benefits. If the domain overlap is low, then constraining the target domain to use the source domain embedding may be harmful relative to training a model from scratch on the target domain. Therefore, we might quantify the source-target distribution shift in terms of how beneficial \lp{} is relative to training a model from scratch (denoted \scr{}):
$$\textit{DomainAffinity}=\text{Acc}_{\textsc{linear}}-\text{Acc}_{\textsc{scratch}}$$
In Figures \ref{fig:intro2} and \ref{fig:ood_metric}, the 19 VTAB-1k target tasks are arranged from low to high by their 
domain affinity to \imgnet~for a pretrained ResNet-50 backbone. The left and right ends of \cref{fig:ood_metric} show examples of the three target domains with the least and most similar distributions, respectively. These examples seem consistent with intuitive notions of distribution shift.

The domain affinity calculated from other pretrained backbones provides a similar ordering.
In \cref{app:metric}, we show that calculating domain affinity using the ViT-B/16 backbone obtains
a high correlation (Spearman, 0.907) with the original scores calculated using the ResNet-50 backbone. Furthermore, we did additional investigations using the 15 representation learning methods presented on the VTAB-leaderboard\footnote{https://google-research.github.io/task\_adaptation/benchmark}, where we calculated median percentage-improvement over scratch training for each task, and similarly observed a high Spearman correlation (0.803).

\vspace{-.5em}
\paragraph{Baselines.} \cref{fig:ood_metric} also presents transfer test accuracy of \lp{}, \ft{}, and \scr{} baselines. Consistent with the literature, \ft{} performs as well or better than the other two baselines.  For in-distribution targets (right side of graph), \lp{} matches \ft{}; for \gls{ood} targets (left side of graph), \lp{} is worse than \ft{}.  With distribution mismatch,  the source network may filter out information available in lower layers because it is not needed for the source task but is essential for the target task. Observing \ft{} performs better than \scr{} even in \gls{ood} tasks, we hypothesize that intermediate features are key for \ft{}, since if learning novel features was possible using limited data, training the network from scratch would work on-par with \gls{ft}. Motivated by this observation, \gls{h2t} probes the intermediate features of a network directly and aims to match the fine-tuning results without modifying the backbone itself.
\begin{figure*}[t]
\centering
\begin{minipage}{.43\textwidth}
\includegraphics[width=\columnwidth]{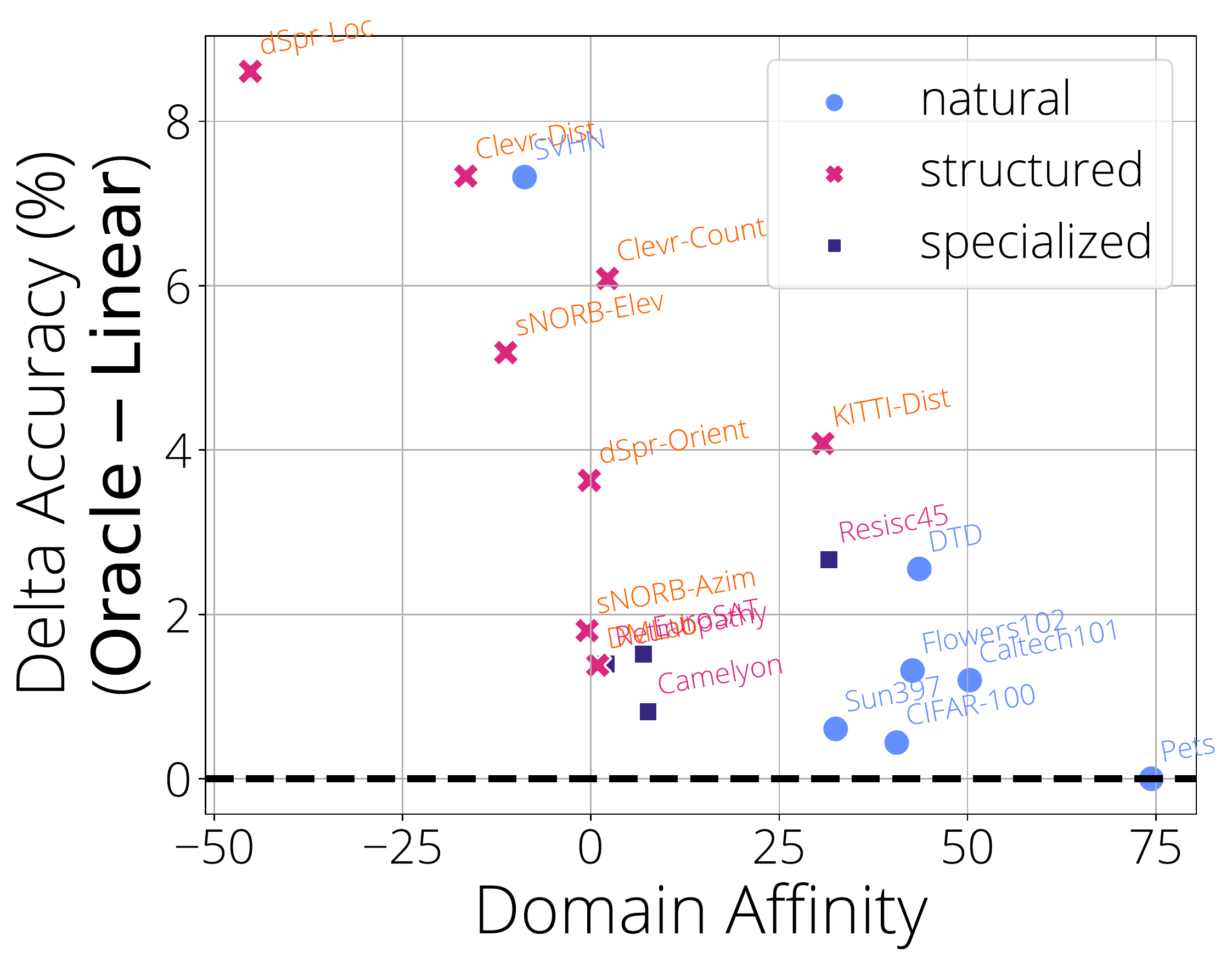}
\end{minipage}%
\begin{minipage}{.5\textwidth}
\centering
\includegraphics[width=.45\columnwidth]{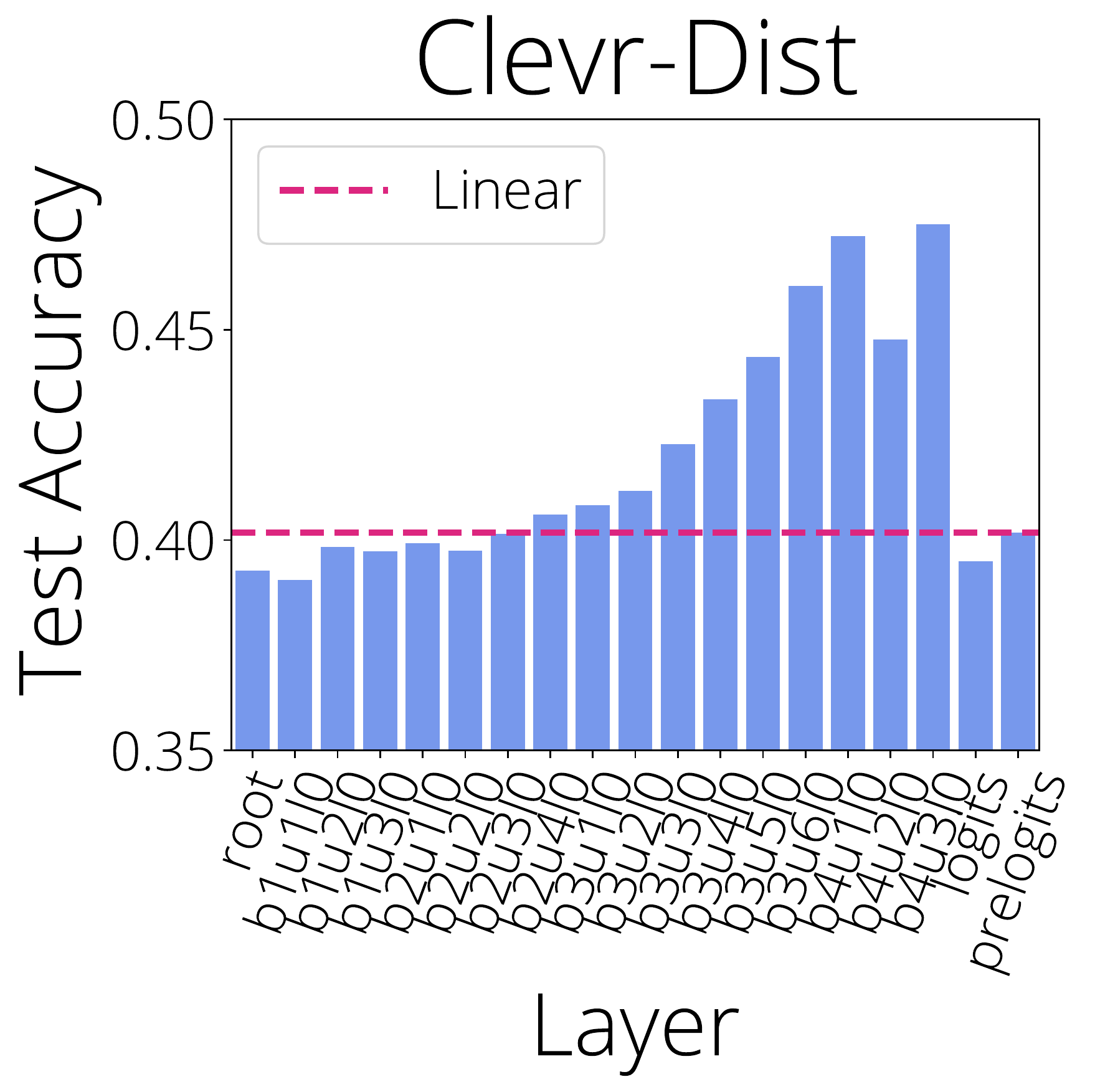}
\includegraphics[width=.45\columnwidth]{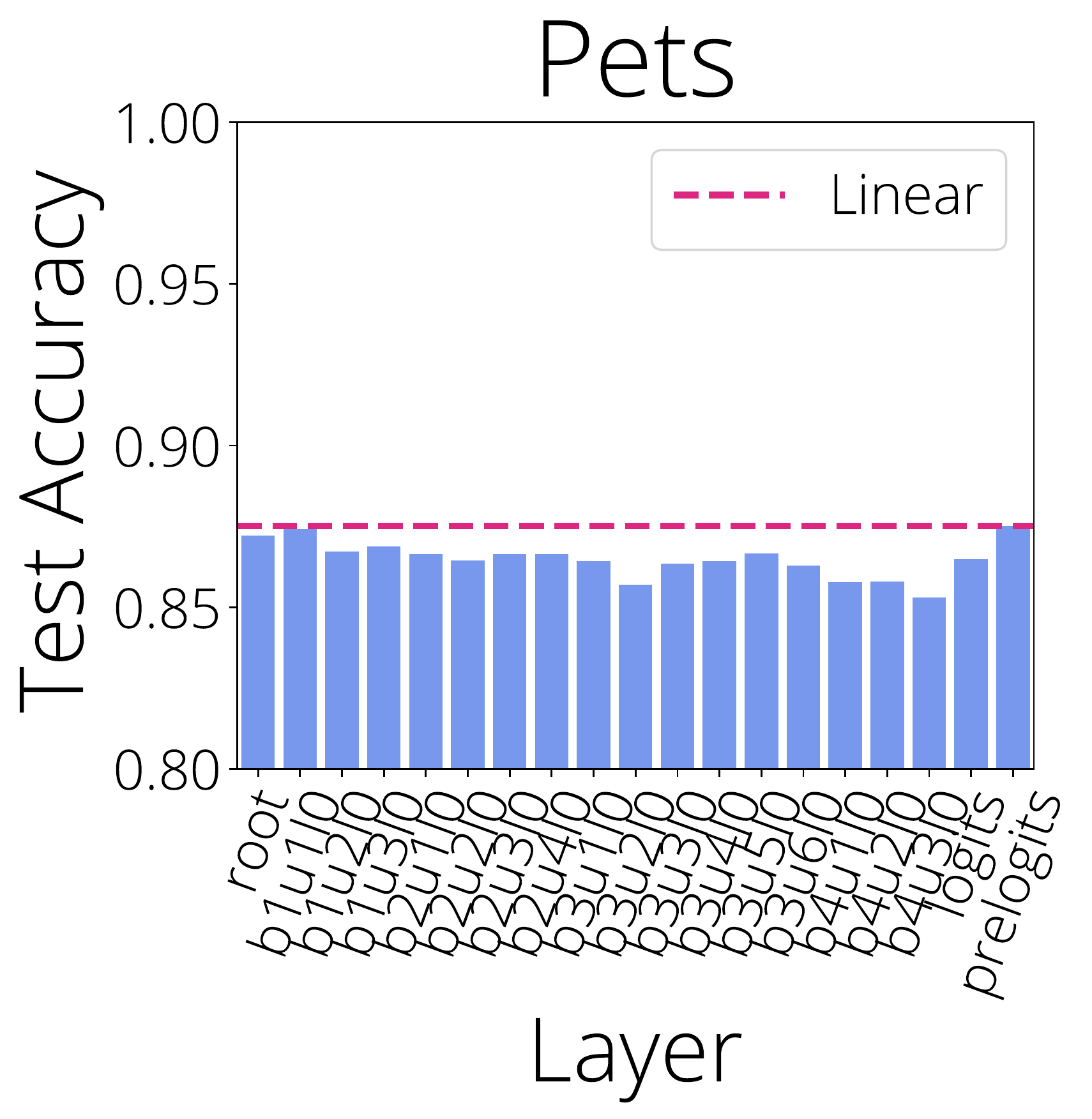}
\small
\begin{tabular}{lrrrr}
\toprule
Method &  Avg &  Spcl &  Strc &  Natr \\
\midrule
Linear   & 56.53 &         80.70 &        36.36 &     65.79 \\
Control & 58.96 &         81.00 &        40.88 &     67.03 \\
Oracle   & 60.15 &         81.52 &        43.44 &     67.03 \\
\bottomrule
\end{tabular}
\end{minipage}%
\caption{\textbf{(left)} Accuracy gains when prelogits are augmented with an additional layer correlates negatively (-0.745, Spearman) with domain affinity. \textbf{(right-top)} effect of using features from intermediate layers for Clevr-Dist (low domain affinity) and Pets (high domain affinity) tasks \textbf{(right-bottom)} Test accuracies of various baselines on VTAB-1k. \textit{Linear} uses only prelogits, \textit{Oracle} averages are obtained by using the layer that gives best generalization (test) for each task. \textit{Control} experiment uses a second feature embedding from a second pretrained network trained  using a different seed. We use ResNet-50 models pretrained on \imgnet.}
\label{fig:ood_oracle}
\end{figure*}

\section{\gls{h2t} Transfer of Pretrained Models}
\subsection{Taylor Approximation of Fine-Tuning}
\label{subsec:taylor}
\citet{Maddox2021FastAW} and \citet{Mu2020GradientsAF} observe that the parameters of a pre-trained backbone change very little during fine-tuning and that a linearized approximation of the fine-tuned model obtains competitive results in transfer learning. Following a similar motivation, we argue that the linearized fine-tuning solution should be well captured by a linear combination of the intermediate activations.

To demonstrate our intuition, consider a multi-layer, fully-connected neural network with input $x$ and scalar output $F(x;w)$, parameterized by weights $w$. We denote the individual elements of $w$ by $w_{ij}$, where the indices reflect the neurons connected such that the activations of neurons are given by $z_j=\sum_i w_{ij}h_i$ and $h_i=f(z_i)$ where $f$ is the activation function used in the network. Then, we can write the fine-tuned neural network, parameterized by optimized parameters $w^*$, using the first-order Taylor approximation:
$$F(x;w^*)\approx F(x;w)+\sum_{i,j} \frac{\partial F(x;w)}{\partial w_{ij}}\Delta w_{ij}$$
where $\Delta w_{ij}=w^*_{ij}-w_{ij}$ reflects the displacement of updated weights during fine-tuning. Expanding the gradient term using the chain rule and rearranging the summations, the linearized solution found by \gls{ft} can be written as a linear combination of the intermediate activations:
\begin{align*} 
F(x;w^*) &\approx F(x;w)+\sum_{i,j} h_i \frac{\partial F(x;w)}{\partial z_{j}}\Delta w_{ij} \\
&\approx F(x;w)+\sum_{i} h_i \sum_{j}\frac{\partial F(x;w)}{\partial z_{j}}\Delta w_{ij} \\ 
&\approx F(x;w)+\sum_{i} h_i c_{i,x}
\end{align*}

Thus, so long as fine-tuning produces small displacements $\Delta w_{ij}$, the \gls{ft} solution for a given input $x$ can be approximated by a linear combination of all features in the network. More broadly, if the coefficients of the most relevant features are robust to input $x$, the \gls{ft} solution has an approximate equivalence to \gls{lp} when trained on features selected from all layers of the network. This conclusion supports \gls{h2t}'s use of intermediate activations as a bridge between \gls{ft} and \gls{lp}.

\subsection{Your Representations are Richer Than You Think}
\label{sec:h2t:2}
In this section, we conduct a simple experiment to demonstrate the potential of using representations from intermediate layers. We concatenate the feature embedding of a pretrained ResNet-50 backbone (features from the penultimate layer) with features from \emph{one} additional layer and train a linear head on top of the concatenated features. When extracting features from convolutional layers, we reduce the dimensionality of the convolutional stack of feature maps using strided average pooling, with a window size chosen so that the resulting number of features is similar to the number of embedding features (2048 for ResNet-50). 

To estimate an upper bound on performance improvement over \lp{} by including a single intermediate layer, we use an oracle to select the layer which yields the largest boost in test performance, separately for each target task. Percentage improvement over \gls{lp} using this \textsc{Oracle} is shown in \cref{fig:ood_oracle}-left. We observe a Spearman correlation of -0.75 between the domain affinity of a target task and the accuracy gain. In accordance with our hypothesis, adding intermediate representations does not improve in-domain generalization because the feature embedding already contains the most useful features. In contrast, generalization on out-of-domain tasks are improved significantly when intermediate features are used.

In \cref{fig:ood_oracle}-top-right, we show test accuracy for two domains as a function of the ResNet-50 layer whose internal representation is used to augment the feature embeddings. Figures for the remaining tasks can be found in \cref{app:oracle_full}. Different tasks on the VTAB-1k benchmark benefit from the inclusion of different layers to obtain the best generalization accuracy, which emphasizes the importance of domain-specific selection of the appropriate set of features for optimal generalization. Overall, the \textsc{Oracle} that selects the layer with best test performance for each task yields an average of 3.5\% improvement on the VTAB-1k benchmark. One possible explanation for the improvement in performance with the augmented representation is simply that it has more degrees of freedom (4096 features instead of 2048). To demonstrate that the improvement is due to inclusion of intermediate layers and not simply due to increased dimensionality, \cref{fig:ood_oracle}-bottom-right compares the \textsc{Oracle} to a \textsc{Control} condition whose representation is matched in dimensionality but formed by concatenating a feature embedding obtained from a second ResNet-50 backbone pretrained on \imgnet{}. Note that this experiment bears similarity to \textit{ensembling} \citep{ZHOU2002ensemble}, which is known to bring significant accuracy gains on its own \citep{Mustafa2020DeepEF}. Using a second backbone doubles the amount of resources required yet falls 1\% shy of \textsc{Oracle} performance, showing the extent to which intermediate representations can be helpful for generalization. 

\subsection{Head2Toe}
\label{sec:h2t:3}
Motivated by our observations in the previous section, we hypothesize that we can attain---or possibly surpass---the performance of \ft{} without modifying the backbone itself by using \lp{} augmented with well-chosen intermediate activations. Our investigation leads us to \gls{h2t}, an efficient transfer learning algorithm based on utilizing intermediate activations using feature selection.
\vspace{-0.5em}
\paragraph{Notation.} Our method applies to any network with any type of layers, but here, for simplicity and without loss of generality, we consider a network with $L$ fully connected layers, each layer receiving input from the layer below:
\begin{align}
  \vz_\ell&=\vh_{\ell-1} \mW_\ell \quad ; \quad  \vh_\ell=f(\vz_\ell) 
\end{align}
where the subscript denotes a layer index, $\vh_0 = \vx$ is the input, $f$ is the activation function, $\mW_\ell$ is the weight matrix of layer $\ell$, and  $\vz_L$ is the logit vector used for classification. 

When transferring a pretrained network to a target task using \lp{}, we discard the last layer of the pretrained network and train a new set of linear weights, $\mW'_{L}$, such that predictions (logits) for the new task are obtained by $\vz'_{L}=\vh_{L-1}\mW'_{L}$.

\begin{figure}[t]
\centering
\includegraphics[width=\columnwidth]{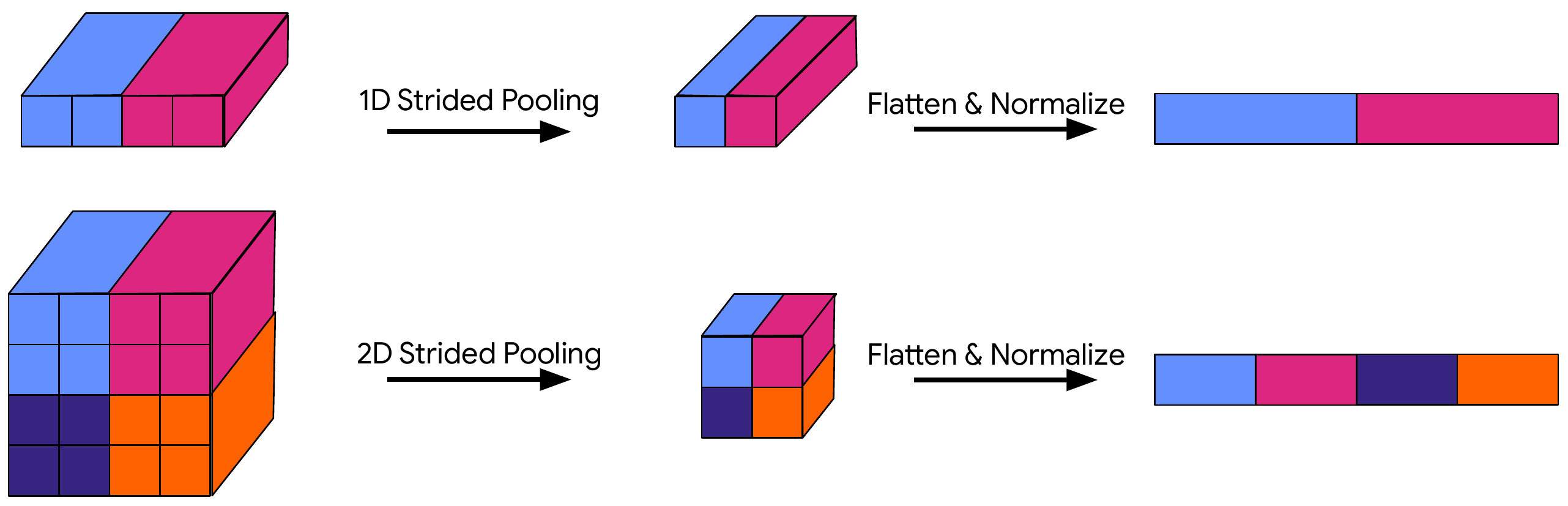}
\caption{We aggregate features over token/spatial dimensions using 1-d/2-d average-pooling and flattening. }
\label{fig:pooling_diag}
\end{figure}

\paragraph{Head2Toe.} Consider a simple scheme that augments the backbone embedding with activations from all layers of the network, such that:
\begin{align}
  \vz'_{L}=\vh_\textit{all}\mW_{all} \quad ; \quad  \vh_\textit{all}= [a_1(\vh_1),..., a_L(\vh_L)]
\end{align}
where $a_\ell(.)$ denotes a fixed function to reduce the dimensionality of the activation vector  and normalize at a given layer $\ell$. Such functions are valuable for network architectures like convolutional networks that generate many intermediate features. Though better aggregation schemes may exist, we simply perform one- or two-dimensional strided average pooling to reduce dimensionality. After aggregation, we normalize features coming from each layer to a unit norm (\cref{fig:pooling_diag}). This scaling preserves the relative magnitude of features within a layer while accounting for inter-layer differences. It works better than normalizing each feature separately or not normalizing at all.  

Even with dimensionality reduction, $\vh_\textit{all}$ can exceed a million elements, and $\mW_{all}$ is underconstrained by the training data, leading to overfitting. Further, $\mW_{all}$ may become so large as to be impractical for deploying this model.\footnote{For example, using a pooling size of 2, ResNet-50 generates 1.7 million features and storing $\mW_{all}$ requires $6.6\times10^8$ parameters (2.6GB for float32) for SUN-397.} We can address these issues by selecting a subset of features before training the target-domain classifier.

\paragraph{Feature selection based on group lasso.} Group lasso \citep{Yuan2006group_lasso} is a popular method for selecting relevant features in multi-task adaptation settings \citep{Argyriou2006,nie2010}. 
When used as a regularizer on a weight matrix $\mW$, the group-lasso regularizer encourages the $\ell_2$ norm of the rows of the matrix to be sparse, and is defined as:
\begin{align}
|\mW|_{2,1}= |\vs|_{1}=\sum_i |s_i| \quad ; \quad  s_i = \sqrt{\sum_j w_{ij}^2} ~~.
\end{align}
To determine which features are most useful for the  task, the linear head is trained with group-lasso regularization on $\mW_{all}$.
In contrast to the approach taken by \citet{Argyriou2006} and \citet{nie2010}, which use costly matrix inversions, we incorporate the regularization as a secondary loss and train with stochastic gradient descent.
Following training, a \emph{relevance score} $s_i$ is computed for each feature $i$. We select a fraction $F$ of the features with the largest relevance and train a new linear head to obtain the final logit mapping. Feature selection alone provides strong regularization, therefore during the final training we do not use any additional regularization.

We make two remarks here. First, because the initial round of training $\mW_\textit{all}$ with the group-lasso regularizer is used only to rank features by importance, the method is robust to the regularization coefficient; it simply needs to be large enough to distinguish the contributions of individual features.
Second, interpreting $s_i$ as the importance of feature $i$ depends on all features having the same dynamic range. This constraint is often satisfied naturally due to the normalization done after aggregation as explained previously. 

\vspace{-.02in}
\paragraph{Selecting $F$.} The fraction $F$ determines the total number of features retained. One would expect the optimal value to depend on the target task. Therefore, we select $F$ for each task separately by cross-validation on the training set. This validation procedure is inexpensive compared to the cost of the initial phase of the algorithm (i.e., training of $\mW_{all}$ to obtain $\vs$) due to the reduced number of features in the second step. Overall, \gls{h2t} together with its validation procedure requires 18\% more operations compared to training $\mW_{all}$ alone (details shared in \cref{app:cost}).

\vspace{-.02in}
\paragraph{Cost of \gls{h2t}.}\gls{h2t}'s use of a fixed backbone means that as we search for features to include, the actual features values are fixed. Consequently, we can calculate them once and re-use as necessary, instead of recalculating at every step, as required for \ft{}. Furthermore, since the backbone is frozen, the storage required for each target task includes only the final output head and the binary mask indicating the features selected. Due to these properties, the cost of \gls{h2t} follows the cost of \gls{lp} closely while being significantly less than \gls{ft}. 

\section{Evaluating Head2Toe}
\label{sec:exps}
We evaluate \gls{h2t} on the VTAB-1k benchmark using two popular vision architectures, ResNet-50 \citep{resnet} and ViT-B/16 \citep{vit}, both pretrained on \imgnet{}. ResNet-50 consists of 50 convolutional layers. To utilize the intermediate convolutional features, we aggregate spatial information using average pooling on non-overlapping patches, as explained in \cref{sec:h2t:3}. We adjust the pooling size to target a fixed dimensionality of the representation. For example, for a feature map of shape $20\times20\times128$ and a \textit{target size} of 512, we average-pool disjoint patches of size $5\times5$, resulting in 4 features per channel and 1024 features in total. This helps us to balance different layers in terms of the number features they contribute to the concatenated embedding. ViT-B/16 consists of 12 multi-headed self-attention layers. When we aggregate the output of the self-attention layers, we perform 1-D average pooling over the patch/token dimension choosing the pooling size to match a target number of features as before. Given that token dimension is permutation invariant, 1-D average pooling is unlikely to be the best choice here and a better aggregation function should provide further gains. We share a detailed list of intermediate representations utilized for each architecture in \cref{app:intermediate_rpr}.

\gls{h2t} selects a subset of features and trains a linear classifier without regularization on top of the selected features. We compare \gls{h2t} with regularization baselines that utilize all features. These baselines are denoted as +All-$\ell_1$, +All-$\ell_2$ and +All-$\ell_{2,1}$ according to the regularizer norm they use.

We perform five-fold cross validation for each task and method in order to pick the best hyperparameters. All methods search over the same learning rates and training steps (two values of each). Methods that leverage intermediate features (i.e., regularization baselines and \gls{h2t}) additionally search over regularization coefficients and the \textit{target size} of the aggregated representation at each layer. The \ft{} baseline searches over 4 hyperparameters; thus the comparison of \gls{h2t}, which searches over 24 values, to fine-tuning might seem unfair. However, this was necessary due to fine-tuning being significantly more costly (about 200x on average) than training a linear classifier on intermediate features. 
We repeat each evaluation using 3 different seeds and report median values and share standard deviations in \cref{app:main_std}. More details on hyperparameter selection and values used are shared in \cref{app:validation}. 

\begin{table*}[t]
\fontsize{7pt}{7pt}\selectfont
\newcolumntype{C}{>{\centering\arraybackslash}X}
\setlength{\tabcolsep}{0pt}
\setlength{\extrarowheight}{5pt}
\renewcommand{\arraystretch}{0.75}
\begin{tabularx}{\linewidth}{p{10pt}p{1.6cm}!{\color{lightgray}\vline} CCCCCCC!{\color{lightgray}\vline}CCCC!{\color{lightgray}\vline}CCCCCCCC!{\color{lightgray}\vline}C}
\toprule
 & & \multicolumn{7}{c}{\underline{Natural}}& \multicolumn{4}{c}{\underline{Specialized}}& \multicolumn{9}{c}{\underline{Structured}}\\
 &
 & \rotatebox{90}{\raisebox{0.5pt}{\tikz\fill[natural] (0,0) circle (.5ex);} CIFAR-100}
 & \rotatebox{90}{\raisebox{0.5pt}{\tikz\fill[natural] (0,0) circle (.5ex);} Caltech101}
 & \rotatebox{90}{\raisebox{0.5pt}{\tikz\fill[natural] (0,0) circle (.5ex);} DTD}
 & \rotatebox{90}{\raisebox{0.5pt}{\tikz\fill[natural] (0,0) circle (.5ex);} Flowers102}
 & \rotatebox{90}{\raisebox{0.5pt}{\tikz\fill[natural] (0,0) circle (.5ex);} Pets}
 & \rotatebox{90}{\raisebox{0.5pt}{\tikz\fill[natural] (0,0) circle (.5ex);} SVHN}
 & \rotatebox{90}{\raisebox{0.5pt}{\tikz\fill[natural] (0,0) circle (.5ex);} Sun397}
 & \rotatebox{90}{\raisebox{0.5pt}{\tikz\fill[specialized] (0,0) circle (.5ex);} Camelyon}
 & \rotatebox{90}{\raisebox{0.5pt}{\tikz\fill[specialized] (0,0) circle (.5ex);} EuroSAT}
 & \rotatebox{90}{\raisebox{0.5pt}{\tikz\fill[specialized] (0,0) circle (.5ex);} Resisc45}
 & \rotatebox{90}{\raisebox{0.5pt}{\tikz\fill[specialized] (0,0) circle (.5ex);} Retinopathy}
 & \rotatebox{90}{\raisebox{0.5pt}{\tikz\fill[structured] (0,0) circle (.5ex);} Clevr-Count}
 & \rotatebox{90}{\raisebox{0.5pt}{\tikz\fill[structured] (0,0) circle (.5ex);} Clevr-Dist}
 & \rotatebox{90}{\raisebox{0.5pt}{\tikz\fill[structured] (0,0) circle (.5ex);} DMLab}
 & \rotatebox{90}{\raisebox{0.5pt}{\tikz\fill[structured] (0,0) circle (.5ex);} KITTI-Dist}
 & \rotatebox{90}{\raisebox{0.5pt}{\tikz\fill[structured] (0,0) circle (.5ex);} dSpr-Loc}
 & \rotatebox{90}{\raisebox{0.5pt}{\tikz\fill[structured] (0,0) circle (.5ex);} dSpr-Ori}
 & \rotatebox{90}{\raisebox{0.5pt}{\tikz\fill[structured] (0,0) circle (.5ex);} sNORB-Azim}
 & \rotatebox{90}{\raisebox{0.5pt}{\tikz\fill[structured] (0,0) circle (.5ex);} sNORB-Elev}
 & \rotatebox{90}{\raisebox{0.5pt}{\tikz\fill[all] (0,0) circle (.5ex);} Mean} \\
\midrule
& \multicolumn{20}{c}{ResNet-50 backbone} \\
& Linear   &           48.5 &           86.0 &           67.8 &           84.8 &           87.4 &           47.5 &           34.4 &  \textbf{83.2} &           92.4 &           73.3 &           73.6 &           39.7 &           39.9 &           36.0 &           66.4 &           40.4 &           37.0 &           19.6 &           25.5 &           57.0 \\
& +All-$\ell_2$        &             44.7 &           87.0 &           67.8 &           84.2 &           86.1 &           81.1 &           31.9 &           82.6 &  \textbf{95.0} &  \textbf{76.5} &           74.5 &           50.0 &           56.3 &           38.3 &           65.5 &           59.7 &           44.5 &           37.5 &           40.0 &           63.3 \\
& +All-$\ell_1$        &    \textbf{50.8} &           88.6 &           67.4 &           84.2 &           87.7 &           84.2 &           34.6 &           80.9 &           94.9 &           75.6 &  \textbf{74.7} &           49.9 &           57.0 &           41.8 &           72.9 &           59.0 &           44.8 &           37.5 &           40.8 &           64.6 \\
& +All-$\ell_{2,1}$       &         49.1 &           86.7 &  \textbf{68.5} &           84.2 &  \textbf{88.0} &  \textbf{84.4} &  \textbf{34.8} &           81.5 &           94.9 &           75.7 &           74.3 &           48.3 &           58.4 &           42.0 &  \textbf{74.4} &           58.8 &           45.2 &           37.8 &           34.4 &           64.3 \\
 & Head2Toe     &            47.1 &  \textbf{88.8} &           67.6 &  \textbf{85.6} &           87.6 &           84.1 &           32.9 &           82.1 &           94.3 &           76.0 &           74.1 &  \textbf{55.3} &  \textbf{59.5} &  \textbf{43.9} &           72.3 &  \textbf{64.9} &  \textbf{51.1} &  \textbf{39.6} &  \textbf{43.1} &  \textbf{65.8} \\
\hline
& Scratch*     &           11.0 &           37.7 &           23.0 &           40.2 &           13.3 &           59.3 &            3.9 &           73.5 &           84.8 &           41.6 &           63.1 &           38.5 &           54.8 &           35.8 &           36.9 &  87.9 &           37.3 &           20.9 &           36.9 &           42.1 \\
& Fine-tuning &         33.2 &           84.6 &           54.5 &  \textbf{85.2} &           79.1 &  \textbf{87.8} &           16.6 &           82.0 &           92.5 &           73.3 &           73.5 &           54.6 &           63.7 &  \textbf{46.3} &           72.1 &  \textbf{94.8} &           47.1 &           35.0 &           33.3 &           63.6 \\
&  Head2Toe-FT     &           16.3 &           87.7 &           63.1 &           84.3 &           66.9 &           82.5 &           24.3 &           82.6 &           11.0 &  \textbf{76.7} &           73.5 &           54.8 &  \textbf{69.1} &           44.7 &           69.2 &           94.2 &  \textbf{51.0} &           33.3 &  \textbf{44.4} &           59.5 \\
& Head2Toe-FT+    &     \textbf{46.9} &  \textbf{88.9} &  \textbf{66.6} &           84.0 &  \textbf{87.3} &           84.4 &  \textbf{32.4} &  \textbf{84.2} &  \textbf{94.4} &           76.7 &  \textbf{74.1} &  \textbf{55.8} &           69.1 &           45.3 &  \textbf{74.7} &           94.4 &           51.0 &  \textbf{39.7} &           42.6 &  \textbf{68.0} \\
\bottomrule
\bottomrule
\end{tabularx}


\fontsize{7pt}{7pt}\selectfont
\newcolumntype{C}{>{\centering\arraybackslash}X}
\setlength{\tabcolsep}{0pt}
\setlength{\extrarowheight}{5pt}
\renewcommand{\arraystretch}{0.75}
\begin{tabularx}{\linewidth}{p{10pt}p{1.6cm}!{\color{lightgray}\vline} CCCCCCC!{\color{lightgray}\vline}CCCC!{\color{lightgray}\vline}CCCCCCCC!{\color{lightgray}\vline}C}
& \multicolumn{20}{c}{ViT-B/16 backbone} \\
& Linear             &       55.0 &           81.0 &           53.6 &           72.1 &           85.3 &           38.7 &           32.3 &           80.1 &           90.8 &           67.2 &           74.0 &           38.5 &           36.2 &           33.5 &           55.7 &           34.0 &           31.3 &           18.2 &           26.3 &           52.8 \\

& +All-$\ell_{2}$          &           57.3 &           87.0 &           64.3 &           82.8 &           84.0 &           75.7 &           32.4 &           82.0 &           94.7 &           79.7 &  \textbf{74.8} &           47.4 &           57.8 &           41.4 &           62.8 &           46.6 &           33.3 &           31.0 &           38.8 &           61.8 \\
& +All-$\ell_{1}$          & 58.4 &  \textbf{87.3} &  \textbf{64.9} &           83.3 &           84.6 &           80.0 &           34.4 &  \textbf{82.3} &  \textbf{95.6} &           79.6 &           73.6 &           47.9 &           57.7 &  \textbf{42.2} &  \textbf{65.1} &           44.5 &           33.4 &           32.4 &           38.4 &           62.4 \\
& +All (Group) &  \textbf{59.6} &           87.1 &           64.9 &           85.2 &  \textbf{85.4} &           79.5 &  \textbf{35.3} &           82.0 &           95.3 &  \textbf{80.6} &           74.2 &           47.9 &           57.8 &           40.7 &           64.9 &           46.7 &  \textbf{33.6} &           31.9 &           39.0 &           62.7 \\
& Head2Toe     &            58.2 &           87.3 &           64.5 &  \textbf{85.9} &           85.4 &  \textbf{82.9} &           35.1 &           81.2 &           95.0 &           79.9 &           74.1 &  \textbf{49.3} &  \textbf{58.4} &           41.6 &           64.4 &  \textbf{53.3} &           32.9 &  \textbf{33.5} &  \textbf{39.4} &  \textbf{63.3} \\
\hline
& Scratch      &           7.6 &           19.1 &           13.1 &           29.6 &            6.7 &           19.4 &            2.3 &           71.0 &           71.0 &           29.3 &           72.0 &           31.6 &           52.5 &           27.2 &           39.1 &           66.1 &           29.7 &           11.7 &           24.1 &           32.8 \\
& Fine-tuning        &      44.3 &           84.5 &           54.1 &           84.7 &           74.7 &  \textbf{87.2} &           26.9 &  \textbf{85.3} &           95.0 &           76.0 &           70.4 &  \textbf{71.5} &           60.5 &           46.9 &  \textbf{72.9} &           74.5 &           38.7 &           28.5 &           23.8 &           63.2 \\

& Head2Toe-FT  &         43.9 &           82.3 &           53.5 &  \textbf{84.9} &           76.7 &           86.5 &           24.5 &           79.9 &           95.9 &           77.5 &  \textbf{74.3} &           68.0 &           70.9 &  \textbf{48.2} &           72.4 &           76.1 &           44.8 &           32.1 &           42.5 &           65.0 \\
& Head2Toe-FT+ &     \textbf{57.3} &  \textbf{87.1} &  \textbf{63.8} &           83.7 &  \textbf{84.8} &           86.8 &  \textbf{35.1} &           80.2 &  \textbf{96.1} &  \textbf{79.9} &           74.1 &           69.9 &  \textbf{71.2} &           47.8 &           72.8 &  \textbf{77.4} &  \textbf{45.9} &  \textbf{33.9} &  \textbf{43.0} &  \textbf{67.9} \\
\bottomrule
\end{tabularx}
\caption{Median test accuracy over 3 seeds on the VTAB-1k benchmark using pretrained backbones. Regularization baselines that use all layers are indicated with the \textit{+All} prefix. "*" indicates results obtained from \citet{Zhai2019TheVT}. Fine-tuning results for ViT-B/16 are obtained using the checkpoints provided by \citet{vit}.}
\label{table:main}
\end{table*}

\subsection{ResNet-50}
The top half of \cref{table:main} presents results on the 19 VTAB-1k target domains when transferring from a pretrained ResNet-50 architecture. On average, \gls{h2t} slightly outperforms all other methods, including \ft{} (see rightmost column of \cref{table:main}). \gls{h2t}, along with the regularization baselines that use intermediate layers, is far superior to \lp{}, indicating the value of the intermediate layers. And \gls{h2t} is superior to the regularization baselines, indicating the value of explicit feature selection. Among the three categories of tasks, \gls{h2t} excels relative to the other methods for the \emph{specialized} category, but does not outperform \ft{} for the \emph{natural} and \emph{structured} categories. 

Does \gls{h2t} select different features for each task? Which layers are used more frequently? In \cref{app:resnet_results}, we show the distribution of features selected across different layers and the amount of intersection among features selected for different tasks and seeds. We observe high variation across tasks, motivating the importance of performing feature selection for each task separately. In addition to the fact that \gls{h2t} outperforms \gls{ft}, it requires only 0.5\% of FLOPs during training on average. Similarly, the cost of storing the adapted model is reduced to 1\% on average. We discuss \gls{h2t}'s computational and storage costs in detail in \cref{app:cost}.
\subsection{ViT-B/16}
Results for ViT-B/16 are shared in the bottom half of \cref{table:main}. As with the ResNet-50 architecture, \gls{h2t} achieves the best accuracy among methods that keep the backbone fixed: \gls{h2t} improves accuracy over \lp{} by about 10\% on average (in absolute performance), and \gls{h2t} outperforms the regularization baselines that include intermediate features but that do not explicitly select features. Similarly to the ResNet-50 experiments, \gls{h2t} matches the performance of \gls{ft}. We share the distribution of features selected over layers in \cref{app:vit_results}. 

\paragraph{\textsc{Head2Toe-FT}.} Our goal in this work is to show that state-of-the-art performance can be obtained efficiently \emph{without} changing the backbone itself. However, we expect to see further gains if the backbone is unfrozen and trained together with the final set of selected features. We performed some initial experiments to demonstrate the potential of such an approach. After selecting features with \gls{h2t}, we fine-tuned the backbone together with the final output layer (\textsc{Head2Toe-FT}) and observed around 2\% increase in accuracy for ViT-B/16. We use the same lightweight validation procedure used for \gls{ft} to pick the learning rate and training steps for the final fine-tuning steps. \textsc{Head2Toe-FT} provides significant gains over \gls{h2t} in the \emph{structured} category, but performs poorly when transferring to some of the \emph{natural} category tasks. Next, we determined whether or not to fine-tune the backbone by examining validation-set accuracy. This procedure, denoted \textsc{Head2Toe-FT+}, provides an additional 3\% improvement and results in 67.9\% accuracy over all VTAB-1k tasks, without any additional training or storage costs compared to \gls{ft}\footnote{\textsc{Head2Toe-FT+} is not a simple max of \textsc{Head2Toe-FT} and \gls{h2t}. This is due to the variance during the re-runs and the mismatch between validation and test accuracies.}.

\begin{figure}[t]
\centering
\includegraphics[width=.66\columnwidth]{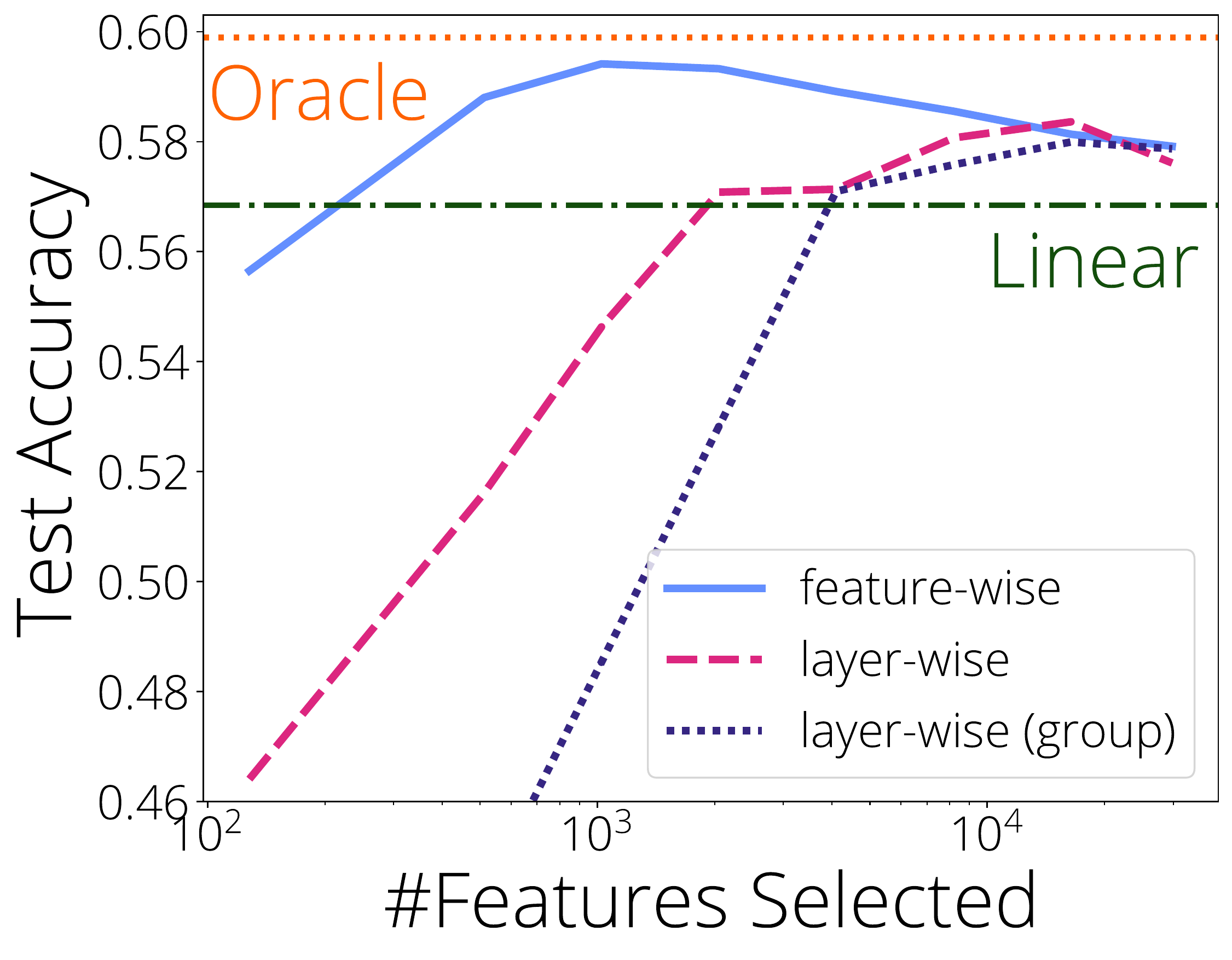}
\includegraphics[width=.33\columnwidth]{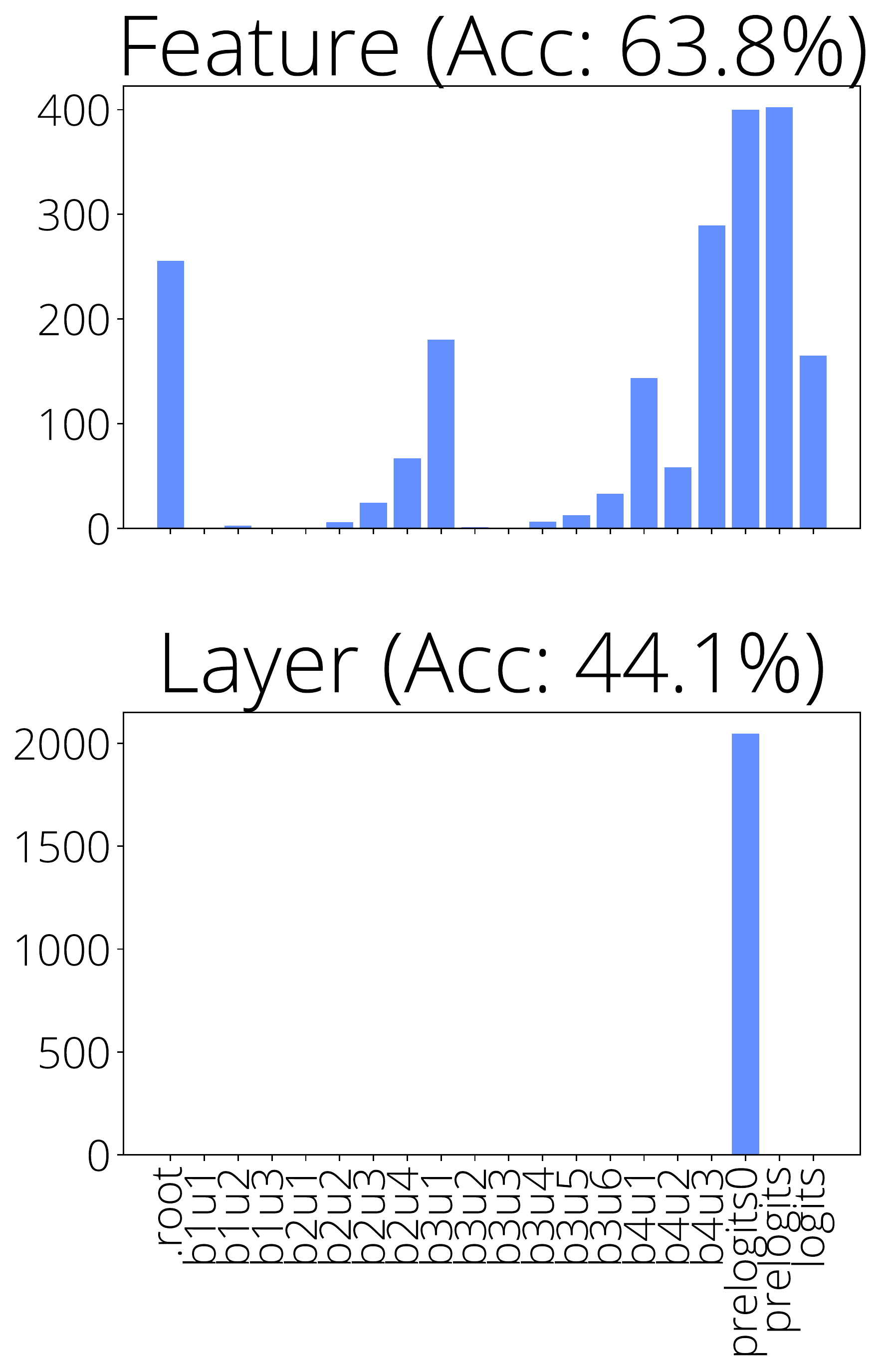}
\caption{\textbf{(left)} Average accuracy over all VTAB tasks as a function of the number of features included. In this experiment, we only use features from the first layer of each of the 18 ResNet blocks and adjust pooling to have around 2048 features for each layer, totalling 29800 features. Results show that selecting layers performs worse than selecting features when adapting to a target domain. \textsc{Oracle} results are explained in \cref{sec:h2t:2}. \textbf{(right)} Distribution of 2048 intermediate features retained from a ResNet-50 when using feature-wise and layer-wise scores on the SVHN transfer task.}
\label{fig:h2t_demo}
\end{figure}

\subsection{Understanding \texorpdfstring{\gls{h2t}}{Head2Toe}}
\gls{h2t} selects individual features from a pre-trained backbone for each task separately and ignores the layer structure of the features. In this section we investigate different parts of the \gls{h2t} algorithm and compare them with some alternatives. \cref{app:resnet_results} and \cref{app:vit_results} include further experiments on the effect of the support set size, choice of intermediate activations and the effectiveness of the relevance scores.
\label{sec:exps:verify}

\begin{figure*}[t]
\centering
\begin{minipage}{.45\textwidth}
\centering
\includegraphics[width=0.95\columnwidth]{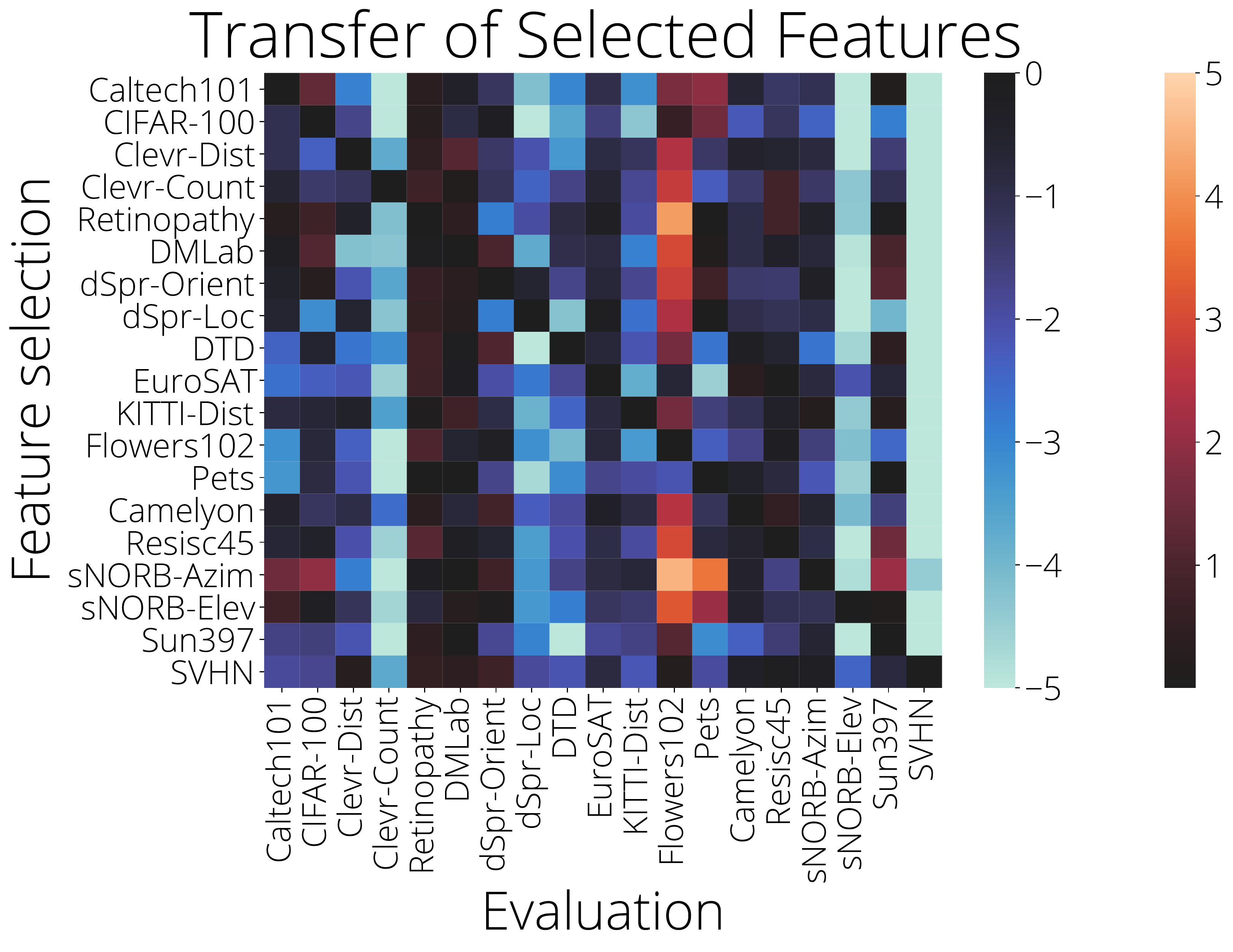}
\end{minipage}%
\begin{minipage}{.35\textwidth}
\includegraphics[width=0.95\columnwidth]{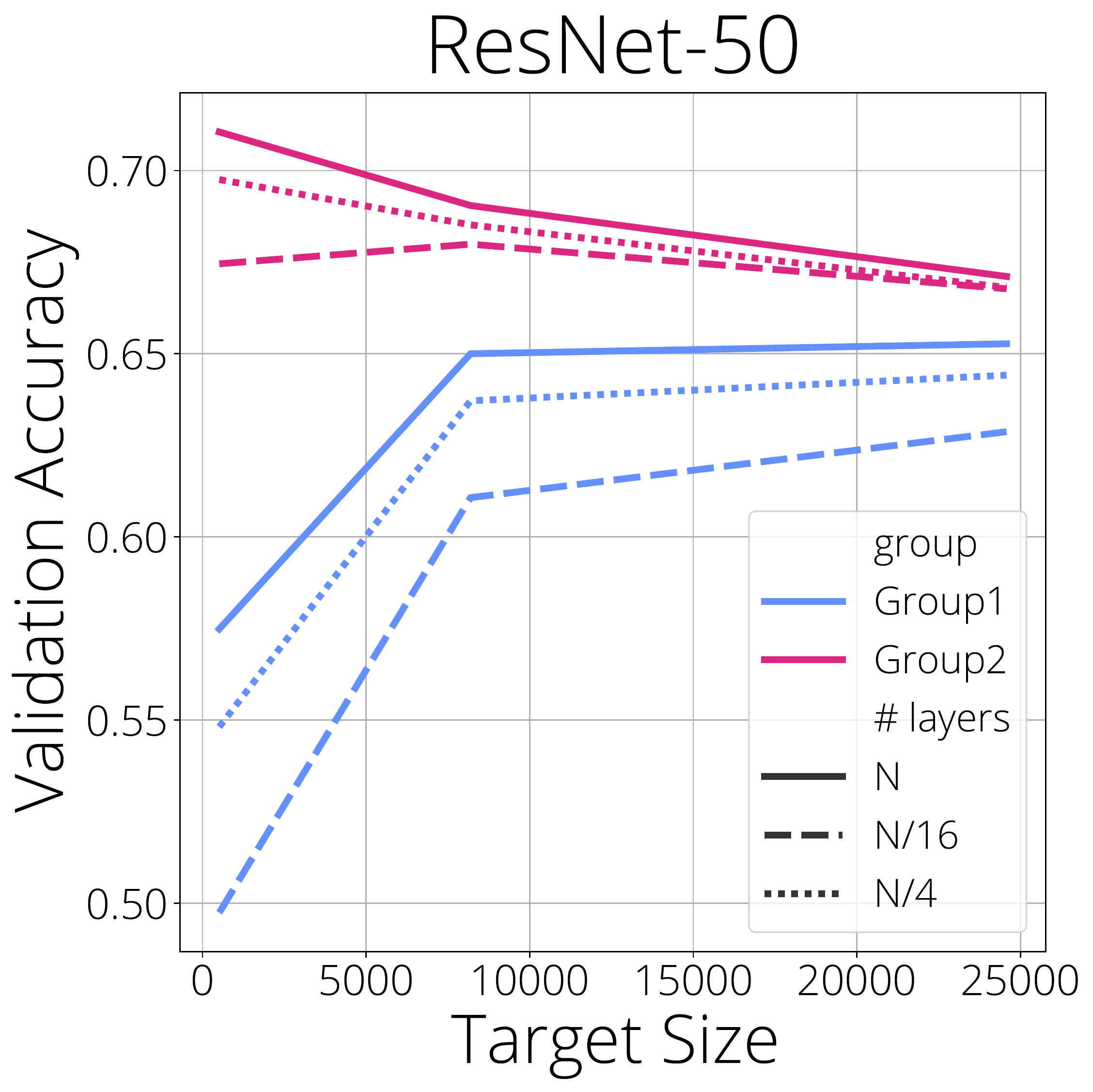}
\end{minipage}%
\begin{minipage}{.25\textwidth}
\fontsize{7pt}{7pt}\selectfont
\setlength{\tabcolsep}{1pt}
\setlength{\extrarowheight}{1pt}
\begin{tabular}{c|c}
Group1 & Group2 \\
\hline
DMLab& CIFAR-100 \\
DTD& Clevr-Count \\
sNORB-Azim& dSpr-Orient \\
SVHN& Retinopathy \\
dSpr-Loc& Resisc45 \\
Pets& EuroSAT \\
sNORB-Elev& Flowers102 \\
& Camelyon \\
& Caltech101 \\
& Clevr-Dist \\
& KITTI-Dist \\
\end{tabular}
\end{minipage}%
\\
\caption{\textbf{(left)} Change in accuracy when features selected for a different task are used for adaptation. Most tasks get their best accuracy when the same task is also used for feature selection. \textbf{(right)} Effect of increasing the number of intermediate features that \gls{h2t} uses from the ResNet-50 backbone. The abscissa of the graph indicates the dimensionality of the representation extracted from each layer of the backbone (\emph{target size}). The tasks are split into two groups (see right side of the figure), which show different behavior. The solid, dashed, and dotted lines indicate the fraction of layers selected for forming the representation used by \gls{h2t}: 1/16, 1/4, and 1, respectively. Sun397 task with all layers and largest target size (24576) failed due to memory issues and thus we omit Sun397 results in these scaling curves.}
\label{fig:understanding}
\end{figure*}

\paragraph{Selecting features or selecting layers?} \gls{h2t} selects individual features independent of the layer in which they are embedded. We compare this \emph{feature-wise} strategy to selecting layers as whole (i.e., selecting all features in a layer or none).  One might expect \emph{layer-wise} selection to be superior because it involves fewer decisions and therefore less opportunity to overfit to the validation set used for selecting the fraction $F$. Further, layer-wise selection may be superior because the relevance of one feature in a layer may indicate the relevance of others. To obtain a layer-wise relevance score, we compute the mean relevance score of all features in a layer and then rank layers by relevance. We also run an alternative layer selection algorithm, in which we group weights originating from the same layer together and select layers using the $\ell_2$ norm of the groups, referred to as \emph{layer-wise (group)}. \cref{fig:h2t_demo}-left compares feature-wise and layer-wise selection methods, matched for number of features retained.  Feature-wise selection performs better than layer-wise selection on average and the best performance is obtained when around 1000 features are kept. \cref{fig:h2t_demo}-right shows the distribution of features selected from each layer by the feature-wise and layer-wise strategies for the SVHN transfer task. We hypothesize that combining features across different layers provide better predictions, while including only the most important features from each layer reduces over-fitting. We share figures for the 19 individual tasks in \cref{app:fraction_full}. 
\vspace{-0.5em}
\paragraph{Transfer across tasks.} In practice, and in the literature, tasks are evaluated in isolation, i.e., the datasets for other tasks are not available. Nonetheless, in \cref{fig:understanding}-left, we investigate how features selected using some task $i$ performs when evaluated on a different task $j$. Each cell in the array represents the average accuracy over three seeds. For each column, we subtract the diagonal term (i.e., self-transfer, $i=j$) to obtain delta accuracy for each task. For most tasks, using a separate task for feature selection hurts performance. Results in Flowers-102 and Pets get better when other tasks like sNORB-Elev are used. Crucially, no single task (rows of \cref{fig:understanding}-left) yields features that are universally optimal, which highlights the importance of doing the feature selection during adaptation, not beforehand. 

\paragraph{Increasing the number of candidate features brings better results.}\gls{h2t} often selects a few thousand features of the over one million features available. The high number of candidate features is critical for obtaining the best performance: in \cref{fig:understanding}-right, we vary the number of intermediate features used for each of our pretrained backbones, indicated by line style. We share scaling curves for the ViT-B/16 backbone in \cref{app:vit_results} and scaling curves for individual tasks in \cref{app:scaling_all}. We observe that including all layers always performs better on average. However when varying the number of target features for each layer, we observed two distinct sub-groups of target tasks that behave differently as the number of features increases, indicated by the red and blue lines. This observation informed our decision to include both small and large target sizes in our validation hyperparameter search. Given the positive slope of the scaling curves, further increasing the number of available features for selection is a promising research direction.

\section{Related Work}
{\bf Transfer Learning} is studied extensively in the literature and used widely in practice. To our knowledge, the utility of combining intermediate layers of a deep neural network is first shown in speech domain \citep{Choi2017TransferLF,Lee2017MultiLevelAM}. Since then many have explored the utility of intermediate representations. \citet{Shor2020TowardsLA,Shor2021UniversalPS} observed earlier layers of an unsupervised speech model to transfer better. ELMo \citep{Peters2018DeepCW} averaged two LSTM embeddings using a learned linear combination (a softmax). \citet{lee2018ood} used intermediate features to improve the detection of \gls{ood} and adversarial inputs. \citet{Tang2020DeepTL} used random projections to reduce the cost of combining different layers, whereas \citet{Adler2020CrossDomainFL} ensembled hebbian learners trained on intermedieate layers and observed larger gains in far target domains. Similar approaches are also used recently to combine multiple backbones \citep{guo2020broader,Lopes2021NoOR} or to improve calibration \citep{khalifa2022improving}. Algorithms that combine layers with all features require embeddings to be same size. Thus, they are most similar to the suboptimal layer-selection baseline shown in \cref{fig:h2t_demo}-left. Most similar to our work is the work of \cite{Dalvi2019WhatIO, Dalvi2020AnalyzingRI}, which proposes a five-step method to select token representations from multiple locations in a pretrained model. They only consider the representation of the [CLS] token after the second MLP of the self-attention block, which makes the feature selection problem significantly smaller and as shown in \cref{app:vit_results} results in sub-optimal transfer. 

Given the ever increasing size and never saturating performance of pretrained models, the importance of reducing the cost of \gls{ft} models is stated in the literature regularly. Methods like feature-wise transformations~\citep{bilen2017universal,dumoulin2018feature}, residual adapters~\citep{Houlsby2019ParamEfficientTL,rebuffi2017learning,Puigcerver2021ScalableTL}, Diff-pruning \citep{Guo2021diffprune} and selective fine-tuning \citep{Guo2019SpotTuneTL,Fu2021LearntoShareAH} are proposed in order to reduce the cost of storing fine-tuned models. However, none of these methods match the simplicity of training (and storing) a linear classifier (i.e. \gls{h2t}) and they can be applied in conjunction with \gls{h2t}. \citet{Teerapittayanon2016BranchyNetFI}, \citet{Kaya2019ShallowDeepNU}, and \citet{Zhou2020BERTLP} studied intermediate representations to reduce "overthinking" and thus provide better early-exit criteria. Similarly \citet{Baldock2021DeepLT} showed a correlation between early classification of a sample and how easy its classification is. Intermedieate features of a pretrained backbone are also used in object detection \citep{Hariharan2015HypercolumnsFO,Bell2016InsideOutsideND,Lin2017FeaturePN} and recently observed to improve the training of generative-adverserial networks \cite{Sauer2021NEURIPS}.  Multiple feature representations are also used by approaches that use multi-domain training as an inductive bias, either without~\citep{dvornik2020selecting,triantafillou2021learning,li2021universal,li2021improving} or with~\citep{liu2020universal} meta-learning. However, in the large-scale setting \gls{ft} remains a top-performing approach to few-shot classification~\citep{dumoulin2021comparing}. 

{\bf Feature selection} approaches can be grouped according to whether labeled data is used---supervised \citep{nie2010} or unsupervised \citep{Ball1965ISODATA,hart1968condensed,he2005laplacian,balin19a,Atashgahi2020QuickAR}---or what high-level approach is taken---filter methods \citep{Langley1994SelectionOR}, wrapper methods \citep{Kohavi1997WrappersFF}, or embedded methods \citep{Yuan2006group_lasso}. Most relevant to our work are embedded supervised methods as they have good scaling properties which is vital in our setting with over a million features. Embedded supervised feature selection methods use a cost function to iteratively refine the subset of features selected and popular approaches include forward selection \citep{Viola2001RapidOD,Borboudakis2019ForwardBackwardSW}, backward selection (pruning) \citep{mozer1989skeletonization,Guyon2004GeneSF} and regularization/feature-ranking based methods \citep{Yuan2006group_lasso,Langley1994SelectionOR,Zhao2010EfficientSF}. Most relevant to our work is \citet{Argyriou2006,nie2010}, both of which uses $\ell_{2,1}$ regularization to select features, however their approach requires matrix inversions which is not practical in our setting. We point interested readers to \citet{Gui2017FeatureSB} and \citet{fs_book} for a detailed discussion. 

\section{Conclusion}
In this work, we introduce \gls{h2t}, an approach that extends linear probing (\lp{}) by selecting the most relevant features among a pretrained network's intermediate representations. We motivate this with a first-order Taylor series approximation to \ft{} and show that the approach greatly improves performance over \lp{} and attains performance competitive with---and in some cases superior to---\ft{} at much lower space and time complexity. Our findings challenge the conventional belief that \ft{} is required to achieve good performance on \gls{ood} tasks. While more work is needed before \gls{h2t} can match the computational efficiency and simplicity of linear probing, our work paves the way for applying new and more efficient feature selection approaches and for experimenting with \gls{h2t} probing in other domains such as regression, video classification, object detection, reinforcement learning, and language modelling.

\section*{Acknowledgments}
We like to thank members of the Google Brain team for their useful feedback. Specifically we like to thank Cristina Vasconcelos, Eleni Triantafillou, Hossein Mobahi, Ross Goroshin for their feedback during the team meetings. We thank Joan Puigcerver, Fabian Pedregosa, Robert Gower, Laura Graesser, Rodolphe Jenatton and Timothy Nguyen for their feedback on the preprint. We thank Lucas Beyer and Xiaohua Zhai for creating the compact table for reporting VTAB results.

\newpage
\bibliography{icml2022}
\bibliographystyle{icml2022}
\newpage
\appendix
\onecolumn
\subsection*{Author Contributions}
\begin{itemize}[leftmargin=*]
\item Utku: Proposed/planned/led the project, wrote the majority of the code, performed most experiments, wrote the intial draft of the paper.
\item Vincent: Participated in weekly meetings for the project, reviewed code, contributed to framing the findings in terms of refuting the hypothesis that a pre-trained network lacks the features required to solve OOD classification tasks, helped with paper writing, helped run evaluations.
\item Hugo: Helped identify and frame the research opportunity, attended regular meetings, contributed to analysis discussions, minor contributions to the writing. Also made this publication possible by pointing out that the paper was over the page-limit just before the submission.
\item Mike: Participated in weekly meetings, confused matters due to his unfamiliarity with the literature, argued that the central idea of the paper was never going to work, contributed to framing research and helped substantially with the writing.
\end{itemize}

\section{Hyperparameter Selection}
\label{app:validation}
We pick hyperparameters for each VTAB task separately by doing a 5-fold cross validation on the training data. For all methods, we chose the learning rate and the total number of training steps using the grid $lr={0.1, 0.01}$ and $steps={500, 5000}$, following the lightweight hyperparameter sweep recommended by the VTAB benchmark \citep{Zhai2019TheVT}. 

For regularization baselines $\ell_1$, $\ell_2$ and $\ell_{2,1}$ we search for regularization coefficients using $(0.00001,0.0001,0.001)$. We include an extra value in this setting in order to account for the overhead introduced by \gls{h2t}.

For \gls{h2t} we choose $\ell_{2,1}$ regularization coefficients from $(0.001, 0.00001)$ and target feature sizes from $(1024,16384,40000)$ for ResNet-50 and $(768,15360,32448)$ for ViT-B/16. After calculating feature scores \gls{h2t} validates the following fractions:  $(0.0005,0.001,0.002,0.005,0.01,0.02,0.05,0.1)$ and thus requires 18\% more operations compared to other regularization baselines. Note that this is because initial training to obtain feature scores $s_i$ is performed once and therefore searching for optimal number of features has a small overhead. Hyper parameters selected by \gls{h2t} for each VTAB task are shared in \cref{table:app:hypers}. Next we explain how this overhead is estimated.
\begin{table}[h]
\centering
\small
\begin{tabular}{lrrrrr|rrrrr}
\toprule
     Dataset &  T &      F &    LR &  Steps &  $\ell_{2,1}$ &  T &      F &    LR &  Steps &  $\ell_{2,1}$  \\
\midrule
&\multicolumn{4}{c}{ResNet-50} &\multicolumn{4}{c}{ViT-B/16}  \\
  Caltech101 &         8192 &  0.010 &  0.01 &   5000 &             0.00001 &          768 &  0.050 &  0.01 &   5000 &             0.00100 \\
   CIFAR-100 &          512 &  0.200 &  0.01 &    500 &             0.00001 &          768 &  0.020 &  0.01 &    500 &             0.00001 \\
  Clevr-Dist &         8192 &  0.001 &  0.01 &    500 &             0.00100 &        15360 &  0.002 &  0.01 &    500 &             0.00100 \\
 Clevr-Count &          512 &  0.005 &  0.10 &   5000 &             0.00100 &          768 &  0.050 &  0.10 &   5000 &             0.00001 \\
 Retinopathy &         8192 &  0.200 &  0.01 &    500 &             0.00001 &          768 &  0.010 &  0.01 &    500 &             0.00100 \\
       DMLab &         8192 &  0.020 &  0.01 &    500 &             0.00001 &        32448 &  0.005 &  0.01 &    500 &             0.00001 \\
 dSpr-Orient &          512 &  0.200 &  0.01 &    500 &             0.00001 &          768 &  0.100 &  0.01 &   5000 &             0.00001 \\
    dSpr-Loc &         8192 &  0.005 &  0.10 &    500 &             0.00100 &        32448 &  0.002 &  0.10 &    500 &             0.00100 \\
         DTD &        24576 &  0.005 &  0.01 &   5000 &             0.00001 &          768 &  0.100 &  0.01 &    500 &             0.00100 \\
     EuroSAT &          512 &  0.100 &  0.01 &    500 &             0.00001 &          768 &  0.100 &  0.01 &    500 &             0.00100 \\
  KITTI-Dist &         8192 &  0.020 &  0.01 &    500 &             0.00001 &        32448 &  0.050 &  0.01 &   5000 &             0.00100 \\
  Flowers102 &          512 &  0.100 &  0.01 &   5000 &             0.00001 &          768 &  0.020 &  0.01 &    500 &             0.00001 \\
        Pets &         8192 &  0.002 &  0.01 &   5000 &             0.00001 &          768 &  0.020 &  0.01 &   5000 &             0.00100 \\
    Camelyon &          512 &  0.020 &  0.10 &    500 &             0.00100 &          768 &  0.100 &  0.01 &    500 &             0.00100 \\
    Resisc45 &         8192 &  0.020 &  0.01 &    500 &             0.00001 &          768 &  0.050 &  0.01 &   5000 &             0.00001 \\
  sNORB-Azim &        24576 &  0.002 &  0.01 &    500 &             0.00001 &        32448 &  0.010 &  0.01 &    500 &             0.00001 \\
  sNORB-Elev &         8192 &  0.050 &  0.01 &    500 &             0.00100 &        15360 &  0.200 &  0.01 &    500 &             0.00100 \\
      Sun397 &          512 &  0.100 &  0.01 &   5000 &             0.00100 &          768 &  0.050 &  0.01 &   5000 &             0.00100 \\
        SVHN &        24576 &  0.005 &  0.01 &    500 &             0.00001 &        32448 &  0.005 &  0.01 &    500 &             0.00001 \\
\bottomrule
\end{tabular}

\caption{Hyper parameters selected for the VTAB-1k benchmark tasks when using pretrained ResNet-50 and ViT-B/16 backbones. \textbf{T}: target sizes of features included from each layer, \textbf{F}: fraction of features kept, \textbf{LR}: learning rate, \textbf{Steps}: Training Steps, \textbf{$\ell_{2,1}$}: regularization coefficient.}
\label{table:app:hypers}
\end{table}

\begin{table}[ht]
\centering
\begin{tabular}{lcccccc}
\toprule
     Dataset &      F &        N &  C & \begin{tabular}{@{}c@{}}FLOPs \\ (vs \gls{ft})\end{tabular} &   \begin{tabular}{@{}c@{}}Size \\ (vs \gls{ft})\end{tabular} &  \begin{tabular}{@{}c@{}}Size \\ (vs \gls{lp})\end{tabular} \\
\midrule
Caltech101 &  0.010 &   467688 &  102 &             0.009675 &            0.020750 &            2.353167 \\
   CIFAR-100 &  0.200 &    30440 &  100 &             0.005792 &            0.025743 &            2.977301 \\
  Clevr-Dist &  0.001 &   467688 &    6 &             0.005747 &            0.000741 &            1.417419 \\
 Clevr-Count &  0.005 &    30440 &    8 &             0.000568 &            0.000092 &            0.132278 \\
 Retinopathy &  0.200 &   467688 &    5 &             0.005657 &            0.020531 &           47.099634 \\
       DMLab &  0.020 &   467688 &    6 &             0.005747 &            0.003011 &            5.756287 \\
 dSpr-Orient &  0.200 &    30440 &   16 &             0.005302 &            0.004183 &            3.001686 \\
    dSpr-Loc &  0.005 &   467688 &   16 &             0.006644 &            0.002212 &            1.587624 \\
         DTD &  0.005 &  1696552 &   47 &             0.015823 &            0.019157 &            4.692396 \\
     EuroSAT &  0.100 &    30440 &   10 &             0.005267 &            0.001336 &            1.532776 \\
  KITTI-Dist &  0.020 &   467688 &    4 &             0.005567 &            0.002215 &            6.350983 \\
  Flowers102 &  0.100 &    30440 &  102 &             0.001117 &            0.013146 &            1.490882 \\
        Pets &  0.002 &   467688 &   37 &             0.003842 &            0.002089 &            0.649417 \\
    Camelyon &  0.020 &    30440 &    2 &             0.005220 &            0.000092 &            0.529114 \\
    Resisc45 &  0.020 &   467688 &   45 &             0.009247 &            0.018474 &            4.725480 \\
  sNORB-Azim &  0.002 &  1696552 &   18 &             0.011069 &            0.004851 &            3.094923 \\
  sNORB-Elev &  0.050 &   467688 &    9 &             0.006016 &            0.009578 &           12.210897 \\
      Sun397 &  0.100 &    30440 &  397 &             0.002839 &            0.049781 &            1.487498 \\
        SVHN &  0.005 &  1696552 &   10 &             0.008464 &            0.005865 &            6.730334 \\
        \hline
        \multicolumn{4}{c}{Average} &       0.006295	&       0.010729 &	5.674742 \\
\bottomrule
\end{tabular}
\caption{Relative cost of \gls{h2t} when compared with \gls{ft} and \gls{lp}. $F$ is the fraction of features kept, $N$ is the total number of features and $C$ is the number of classes. On average \gls{h2t} requires 0.5\% of the FLOPs required by \gls{ft} during the adaptation. Cost of storing each adapted model is also small: requiring only 1\% of the \gls{ft} and only 5.7x more than \gls{lp}. See main text for details on how the numbers are calculated.} 
\label{table:app:resnet_cost}
\end{table}
\section{Cost of \texorpdfstring{\gls{h2t}}{Head2Toe}}
\label{app:cost}

We evaluate different values of $F$ and pick the value with best validation performance. Cost of \gls{h2t} consists of three parts: (1) $C_{I}$: Cost of calculating the representations using the pretrained backbone. (2) $C_{all}$: cost of training the initial head $\mW_{all}$ (in order to obtain $s_i$'s) (3) $\sum_f C_{F=f}$: total cost of validating different values of $F$. Cost of validating a fraction value $f$, assuming equal number of training steps, is equal to $C_f = C_{all} *f$. Therefore relative cost of searching for $F$ is equal to the sum of fractions validated (in comparison to the initial training of ($\vs$)).

In \cref{table:app:resnet_cost}, we compare cost of running \gls{h2t} adaptation with \gls{ft}. \gls{h2t} uses the backbone once in order to calculate the representations and then trains the $\mW_{all}$, whereas \gls{ft} requires a forward pass on the backbone at each step. Therefore we calculate the cost of fine-tuning for $t$ steps as $C_I\cdot t$. Similarly, cost of \gls{h2t} is calculated as $C_I+C_{all}\cdot t$. The overall relative cost of $C_{all}$ increases with number of classes $C$ and number of features considered $N$. As shown in \cref{table:main}-top, \gls{h2t} obtains better results than \gls{ft}, yet it requires 0.5\% of the FLOPs needed on average.

After adaptation, all methods require roughly same number of FLOPs for inference due to all methods using the same backbone. The cost of storing models for each task becomes important when the same pre-trained backbone is used for many different tasks. In \cref{table:app:resnet_cost} we compare the cost of storing different models found by different methods. A fine-tuned model requires all weights to be stored which has the same cost as storing the original network, whereas \gls{lp} and \gls{h2t} requires storing only the output head: $W_{linear}$. \gls{h2t} also requires to store the indices of the features selected using a bitmap. Even though \gls{h2t} considers many more features during adaptation, it selects a small subset of the features and thus requires a much smaller final classifier (on average 1\% of the \gls{ft}). Note that hyper parameters are selected to maximize accuracy, not the size of the final model. We expect to see greater savings with an efficiency oriented hyperparameter selection.

\section{Details of Intermediate Representations Included}
\label{app:intermediate_rpr}
\paragraph{ResNet-50} has 5 stages. First stage includes a single convolutional layer (root) followed by pooling. We include features after pooling. Remaining stages include 3,4,6 and 3 bottleneck units (v2) each. Bottleneck units start with normalization and activation and has 3 convolutional layers. We includes features right after the activation function is applied, resulting in 3 intermediate feature sets per unit. Output after 5 stages are average-pooled and passed to the output head. We include features before and after pooling. with the output of the final layer (logits), total number of locations where features are extracted makes 52.

\paragraph{ViT-B/16} model consists of multiple encoder units. Each encoder unit consists of a self attention module followed by 2 MLPs. For each of these units, we include features (1) after layer-norm (but before self-attention) (2) features after self-attention (3-4) features after MLP layers (and after gelu activation function). Additionally we use patched and embedded image (i.e. tokenized image input), pre-logits (final CLS-embedding) and logits. 

\newpage

\section{Standard Deviations for \texorpdfstring{\cref{table:main}}{Table 1}}
\label{app:main_std}
In \cref{app:table:main_std}, we share the standard deviations for the median test accuracies presented in \cref{table:main}. On average we observe \gls{lp} obtains lower variation as expected, due to the limited adaptation and convex nature of the problem. Head2Toe seem to have similar (or less) variation then the other regularization baselines that use all features.
\begin{table}[h]
\fontsize{7pt}{7pt}\selectfont
\newcolumntype{C}{>{\centering\arraybackslash}X}
\setlength{\tabcolsep}{0pt}
\setlength{\extrarowheight}{5pt}
\renewcommand{\arraystretch}{0.75}
\begin{tabularx}{\linewidth}{p{10pt}p{1.6cm}!{\color{lightgray}\vline} CCCCCCC!{\color{lightgray}\vline}CCCC!{\color{lightgray}\vline}CCCCCCCC!{\color{lightgray}\vline}C}
\toprule
 & & \multicolumn{7}{c}{\underline{Natural}}& \multicolumn{4}{c}{\underline{Specialized}}& \multicolumn{9}{c}{\underline{Structured}}\\
 &
 & \rotatebox{90}{\raisebox{0.5pt}{\tikz\fill[natural] (0,0) circle (.5ex);} CIFAR-100}
 & \rotatebox{90}{\raisebox{0.5pt}{\tikz\fill[natural] (0,0) circle (.5ex);} Caltech101}
 & \rotatebox{90}{\raisebox{0.5pt}{\tikz\fill[natural] (0,0) circle (.5ex);} DTD}
 & \rotatebox{90}{\raisebox{0.5pt}{\tikz\fill[natural] (0,0) circle (.5ex);} Flowers102}
 & \rotatebox{90}{\raisebox{0.5pt}{\tikz\fill[natural] (0,0) circle (.5ex);} Pets}
 & \rotatebox{90}{\raisebox{0.5pt}{\tikz\fill[natural] (0,0) circle (.5ex);} SVHN}
 & \rotatebox{90}{\raisebox{0.5pt}{\tikz\fill[natural] (0,0) circle (.5ex);} Sun397}
 & \rotatebox{90}{\raisebox{0.5pt}{\tikz\fill[specialized] (0,0) circle (.5ex);} Camelyon}
 & \rotatebox{90}{\raisebox{0.5pt}{\tikz\fill[specialized] (0,0) circle (.5ex);} EuroSAT}
 & \rotatebox{90}{\raisebox{0.5pt}{\tikz\fill[specialized] (0,0) circle (.5ex);} Resisc45}
 & \rotatebox{90}{\raisebox{0.5pt}{\tikz\fill[specialized] (0,0) circle (.5ex);} Retinopathy}
 & \rotatebox{90}{\raisebox{0.5pt}{\tikz\fill[structured] (0,0) circle (.5ex);} Clevr-Count}
 & \rotatebox{90}{\raisebox{0.5pt}{\tikz\fill[structured] (0,0) circle (.5ex);} Clevr-Dist}
 & \rotatebox{90}{\raisebox{0.5pt}{\tikz\fill[structured] (0,0) circle (.5ex);} DMLab}
 & \rotatebox{90}{\raisebox{0.5pt}{\tikz\fill[structured] (0,0) circle (.5ex);} KITTI-Dist}
 & \rotatebox{90}{\raisebox{0.5pt}{\tikz\fill[structured] (0,0) circle (.5ex);} dSpr-Loc}
 & \rotatebox{90}{\raisebox{0.5pt}{\tikz\fill[structured] (0,0) circle (.5ex);} dSpr-Ori}
 & \rotatebox{90}{\raisebox{0.5pt}{\tikz\fill[structured] (0,0) circle (.5ex);} sNORB-Azim}
 & \rotatebox{90}{\raisebox{0.5pt}{\tikz\fill[structured] (0,0) circle (.5ex);} sNORB-Elev}
 & \rotatebox{90}{\raisebox{0.5pt}{\tikz\fill[all] (0,0) circle (.5ex);} Mean} \\
\midrule
&Linear   &      0.09 &       0.08 &  0.14 &       0.06 &  0.08 &  0.17 &   0.06 &     0.06 &    0.03 &     0.12 &        0.08 &        0.42 &        0.1 &  0.03 &       0.21 &     0.14 &        0.07 &        0.0 &       0.21 &  0.11 \\
&+All-$\ell_2$   &      0.09 &       0.78 &  0.44 &       0.29 &  0.22 &  0.02 &   0.04 &     0.02 &    0.06 &     0.05 &        0.02 &        0.09 &        0.2 &   0.1 &       1.31 &     0.32 &        0.08 &       0.03 &       0.35 &  0.24 \\
& +All-$\ell_1$   &      0.14 &       0.11 &  0.11 &       0.11 &  0.11 &  0.12 &   0.03 &     0.06 &    0.04 &      0.2 &        0.02 &        0.47 &       0.33 &  0.18 &        0.2 &     0.34 &        0.07 &       0.06 &       0.44 &  0.17 \\
& +All-$\ell_{2,1}$  &      0.09 &        1.0 &  0.13 &        0.1 &   0.1 &  0.18 &   0.23 &     0.15 &    0.07 &     0.09 &        0.05 &        0.03 &       0.08 &  0.11 &       0.41 &     0.51 &        0.24 &       0.16 &       0.07 &   0.2 \\
&Head2Toe &      0.14 &       0.25 &  0.08 &       0.08 &  0.24 &  0.24 &   0.16 &     0.23 &    0.06 &     0.06 &        0.03 &        0.18 &       0.23 &  0.13 &       0.43 &      0.3 &        0.06 &        0.4 &       0.08 &  0.18 \\
\hline
& Fine-tuning&      0.53 &       1.05 &  0.18 &       0.62 &  0.16 &  0.24 &   0.86 &      2.2 &    0.52 &     0.57 &        0.37 &        2.39 &       2.49 &  0.43 &       1.06 &     0.54 &        0.33 &       0.39 &       0.73 &  0.83 \\
& Head2Toe-FT     &      0.23 &      50.52 &  0.47 &       0.49 &  0.74 &  0.05 &  13.78 &     0.39 &     0.0 &     0.64 &         0.0 &        3.02 &       1.53 &   0.9 &       1.11 &     0.83 &        0.43 &       0.46 &       0.76 &  4.02 \\
& Head2Toe-FT+    &      0.29 &       0.15 &  0.45 &       0.08 &  0.28 &  0.25 &   0.12 &     1.61 &    0.08 &     0.15 &        0.18 &        0.27 &       0.67 &   1.0 &       0.76 &     0.96 &         0.1 &       0.38 &       2.09 &  0.52 \\
\bottomrule
\end{tabularx}


\fontsize{7pt}{7pt}\selectfont
\newcolumntype{C}{>{\centering\arraybackslash}X}
\setlength{\tabcolsep}{0pt}
\setlength{\extrarowheight}{5pt}
\renewcommand{\arraystretch}{0.75}
\begin{tabularx}{\linewidth}{p{10pt}p{1.6cm}!{\color{lightgray}\vline} CCCCCCC!{\color{lightgray}\vline}CCCC!{\color{lightgray}\vline}CCCCCCCC!{\color{lightgray}\vline}C}
\toprule
 & & \multicolumn{7}{c}{\underline{Natural}}& \multicolumn{4}{c}{\underline{Specialized}}& \multicolumn{9}{c}{\underline{Structured}}\\
 &
 & \rotatebox{90}{\raisebox{0.5pt}{\tikz\fill[natural] (0,0) circle (.5ex);} CIFAR-100}
 & \rotatebox{90}{\raisebox{0.5pt}{\tikz\fill[natural] (0,0) circle (.5ex);} Caltech101}
 & \rotatebox{90}{\raisebox{0.5pt}{\tikz\fill[natural] (0,0) circle (.5ex);} DTD}
 & \rotatebox{90}{\raisebox{0.5pt}{\tikz\fill[natural] (0,0) circle (.5ex);} Flowers102}
 & \rotatebox{90}{\raisebox{0.5pt}{\tikz\fill[natural] (0,0) circle (.5ex);} Pets}
 & \rotatebox{90}{\raisebox{0.5pt}{\tikz\fill[natural] (0,0) circle (.5ex);} SVHN}
 & \rotatebox{90}{\raisebox{0.5pt}{\tikz\fill[natural] (0,0) circle (.5ex);} Sun397}
 & \rotatebox{90}{\raisebox{0.5pt}{\tikz\fill[specialized] (0,0) circle (.5ex);} Camelyon}
 & \rotatebox{90}{\raisebox{0.5pt}{\tikz\fill[specialized] (0,0) circle (.5ex);} EuroSAT}
 & \rotatebox{90}{\raisebox{0.5pt}{\tikz\fill[specialized] (0,0) circle (.5ex);} Resisc45}
 & \rotatebox{90}{\raisebox{0.5pt}{\tikz\fill[specialized] (0,0) circle (.5ex);} Retinopathy}
 & \rotatebox{90}{\raisebox{0.5pt}{\tikz\fill[structured] (0,0) circle (.5ex);} Clevr-Count}
 & \rotatebox{90}{\raisebox{0.5pt}{\tikz\fill[structured] (0,0) circle (.5ex);} Clevr-Dist}
 & \rotatebox{90}{\raisebox{0.5pt}{\tikz\fill[structured] (0,0) circle (.5ex);} DMLab}
 & \rotatebox{90}{\raisebox{0.5pt}{\tikz\fill[structured] (0,0) circle (.5ex);} KITTI-Dist}
 & \rotatebox{90}{\raisebox{0.5pt}{\tikz\fill[structured] (0,0) circle (.5ex);} dSpr-Loc}
 & \rotatebox{90}{\raisebox{0.5pt}{\tikz\fill[structured] (0,0) circle (.5ex);} dSpr-Ori}
 & \rotatebox{90}{\raisebox{0.5pt}{\tikz\fill[structured] (0,0) circle (.5ex);} sNORB-Azim}
 & \rotatebox{90}{\raisebox{0.5pt}{\tikz\fill[structured] (0,0) circle (.5ex);} sNORB-Elev}
 & \rotatebox{90}{\raisebox{0.5pt}{\tikz\fill[all] (0,0) circle (.5ex);} Mean} \\
\midrule
&Linear       &       0.1 &       0.23 &  0.19 &       0.16 &  0.06 &  0.06 &   0.08 &     0.09 &    0.08 &     0.06 &         0.0 &        0.06 &       0.02 &  0.07 &       0.92 &     0.21 &        0.08 &       0.08 &       0.03 &  0.14 \\
&+All-$\ell_2$    &      0.13 &       0.17 &   0.0 &       0.66 &  0.36 &  0.04 &   0.56 &     0.02 &    0.59 &     0.01 &        0.04 &        0.23 &       0.13 &  0.12 &       1.24 &     0.28 &        0.09 &       0.84 &        0.2 &   0.3 \\
&+All-$\ell_1$    &      0.06 &       0.11 &  0.15 &       0.08 &  0.19 &  0.31 &   0.08 &     0.08 &    0.06 &     0.12 &        0.05 &        0.28 &       0.17 &  0.23 &       0.87 &     0.34 &        0.07 &       0.22 &       0.28 &   0.2 \\
&+All-$\ell_{2,1}$ &      1.55 &       0.03 &  0.12 &       0.04 &  0.06 &  0.41 &   0.13 &     0.18 &    0.08 &      0.1 &        0.04 &        0.09 &       0.13 &  0.15 &       0.87 &     0.66 &         0.1 &       0.23 &       0.22 &  0.27 \\
&Head2Toe     &      0.29 &       0.16 &  0.26 &        0.5 &  0.19 &  0.14 &   0.07 &     0.13 &    0.04 &     0.09 &        0.09 &        0.04 &       0.44 &  0.14 &       1.34 &     0.21 &        0.01 &       0.31 &       0.08 &  0.24 \\
\hline
& Scratch      &      0.25 &       0.29 &  0.43 &       0.68 &  0.16 &  1.09 &    0.2 &     0.57 &    0.79 &     1.03 &        1.63 &        0.13 &       0.94 &  0.52 &        1.8 &     4.73 &        3.17 &       0.61 &       1.39 &  1.08 \\
& Fine-tuning     &      1.71 &       0.89 &  0.13 &       0.83 &  0.81 &  3.41 &   0.33 &     1.59 &    0.45 &     0.43 &         1.1 &        1.99 &       0.47 &  0.25 &       1.24 &     0.98 &        3.52 &       1.33 &        1.8 &  1.22 \\
&Head2Toe-FT  &      0.68 &       0.39 &  1.05 &       0.05 &  0.65 &  0.45 &    0.3 &     0.24 &     0.1 &     0.76 &        0.29 &        1.78 &       1.33 &  0.21 &       0.81 &     2.69 &        0.45 &       0.64 &       1.65 &  0.76 \\
&Head2Toe-FT+ &      0.25 &       0.08 &  0.16 &       0.24 &  0.28 &  0.21 &   0.06 &     0.55 &    0.13 &     0.14 &        0.03 &        1.84 &       1.03 &   0.5 &       1.29 &     2.32 &        1.98 &       0.19 &       0.39 &  0.61 \\

\bottomrule
\end{tabularx}

\caption{Standard deviation of test accuracy over 3 seeds on the VTAB-1k benchmark using pretrained (\textbf{top}) ResNet-50 and (\textbf{bottom}) ViT-B/16 backbones. The mean column averages the standard deviations for each dataset.}
\label{app:table:main_std}
\end{table}

\section{Details of Datasets used in VTAB-1k bechmark}
\cref{tab:tasks} include datasets used in VTAB-1k benchmark.
\begin{table}[h]
\centering
\small
\begin{tabular}{lrl}
\toprule
\bf Dataset & \bf Classes & \bf Reference \\
\toprule
Caltech101 & 102 & \cite{li2006one} \\
CIFAR-100  & 100 & \cite{cifar10} \\
DTD & 47 & \cite{cimpoi14describing} \\
Flowers102 & 102 & \cite{Nilsback08} \\
Pets & 37 & \cite{parkhi12a} \\
Sun397 & 397 & \cite{xiao2010sun} \\
SVHN & 10 & \cite{netzer2011reading} \\
EuroSAT & 10 & \cite{helber2017eurosat} \\
Resisc45 & 45 & \cite{cheng2017remote} \\
Patch Camelyon & 2 & \cite{veeling2018rotation} \\
Retinopathy & 5 & \cite{kaggle-diabetic-retinopathy} \\
Clevr  & 8 & \cite{johnson2017clevr}\\
dSprites & 16 & \cite{dsprites17} \\
SmallNORB & 18 & \cite{lecun2004learning} \\
DMLab & 6 & \cite{beattie2016deepmind} \\
KITTI/distance &  4 & \cite{Geiger2013IJRR} \\
\bottomrule
\end{tabular}
\caption{Datasets used in VTAB-1k benchmark.}
\label{tab:tasks}
\end{table}

\newpage

\section{Domain Affinity Metric}
\label{app:metric}
Domain affinity metric aims to capture the similarity between two supervised learning task. If two tasks are similar, we expect the representations learned in one task to transfer better using a simple linear probe on the last layer compared to training the network from scratch in a low-data regime. Thus we define our domain affinity as the difference between linear probe and scratch accuracies. Here we demonstrate the robustness of the domain affinity metric to different backbone architectures and pretraining algorithms.

We expect the domain affinity calculated using different pretrained backbones to provide similar orderings. We demonstrate this in \cref{app:fig:singlefeat}-left, where we calculate domain affinity using the ViT-B/16 backbone and observe a high correlation (Spearman, 0.907) with the original scores calculated using the ResNet-50 backbone. Furthermore we did additional investigations using the 15 representation learning methods presented on the VTAB-leaderboard\footnote{https://google-research.github.io/task\_adaptation/benchmark}, where we calculated median percentage-improvement over scratch training for each task, and similarly observed a high Spearman correlation (0.803).

\section{Additional Plots for Experiments using Single Additional Layer}
\label{app:oracle_full}

\begin{figure*}[t]
\centering
\begin{minipage}{.45\textwidth}
\includegraphics[width=0.95\columnwidth]{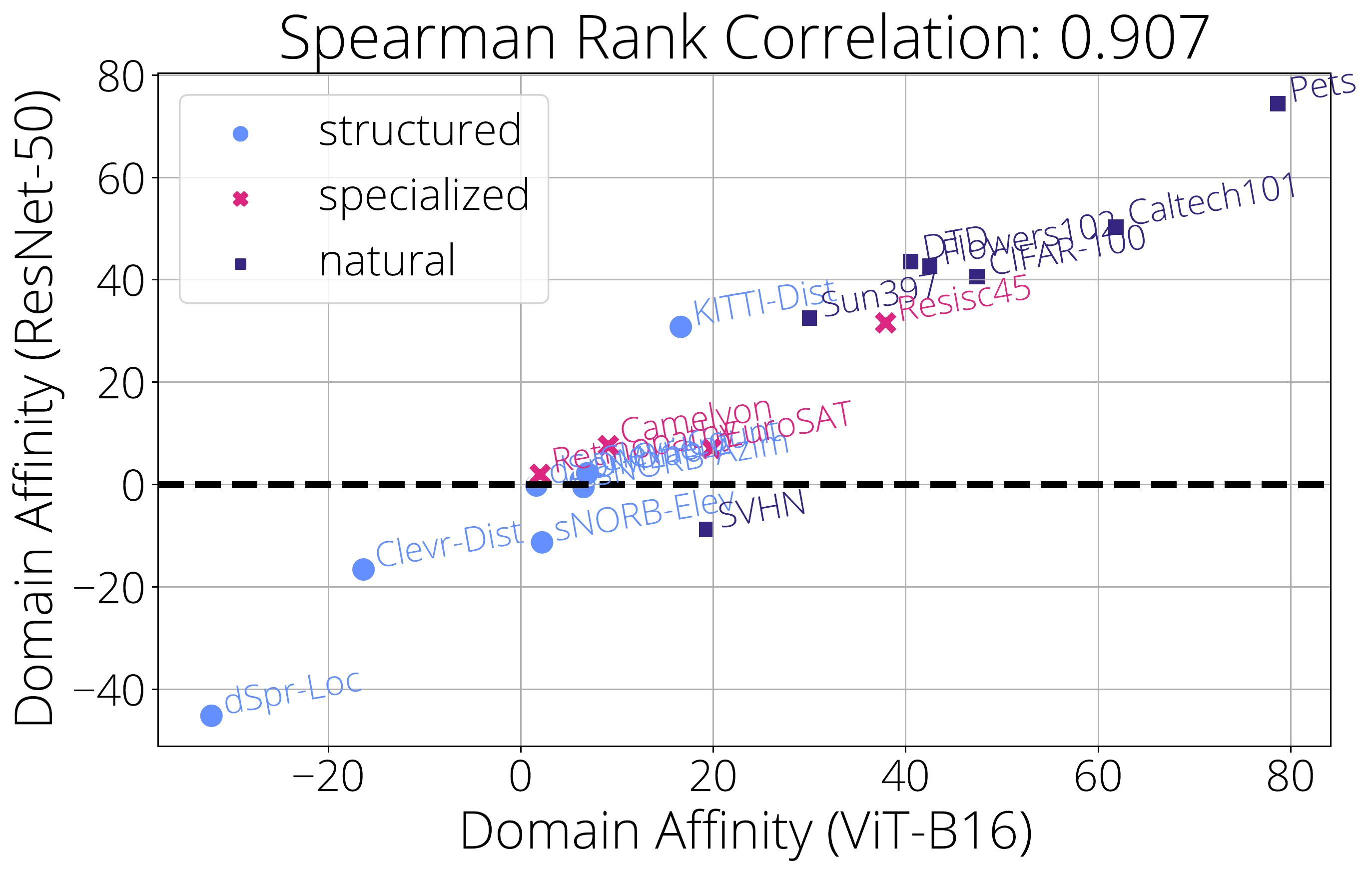}
\end{minipage}%
\begin{minipage}{.45\textwidth}
\includegraphics[width=0.95\columnwidth]{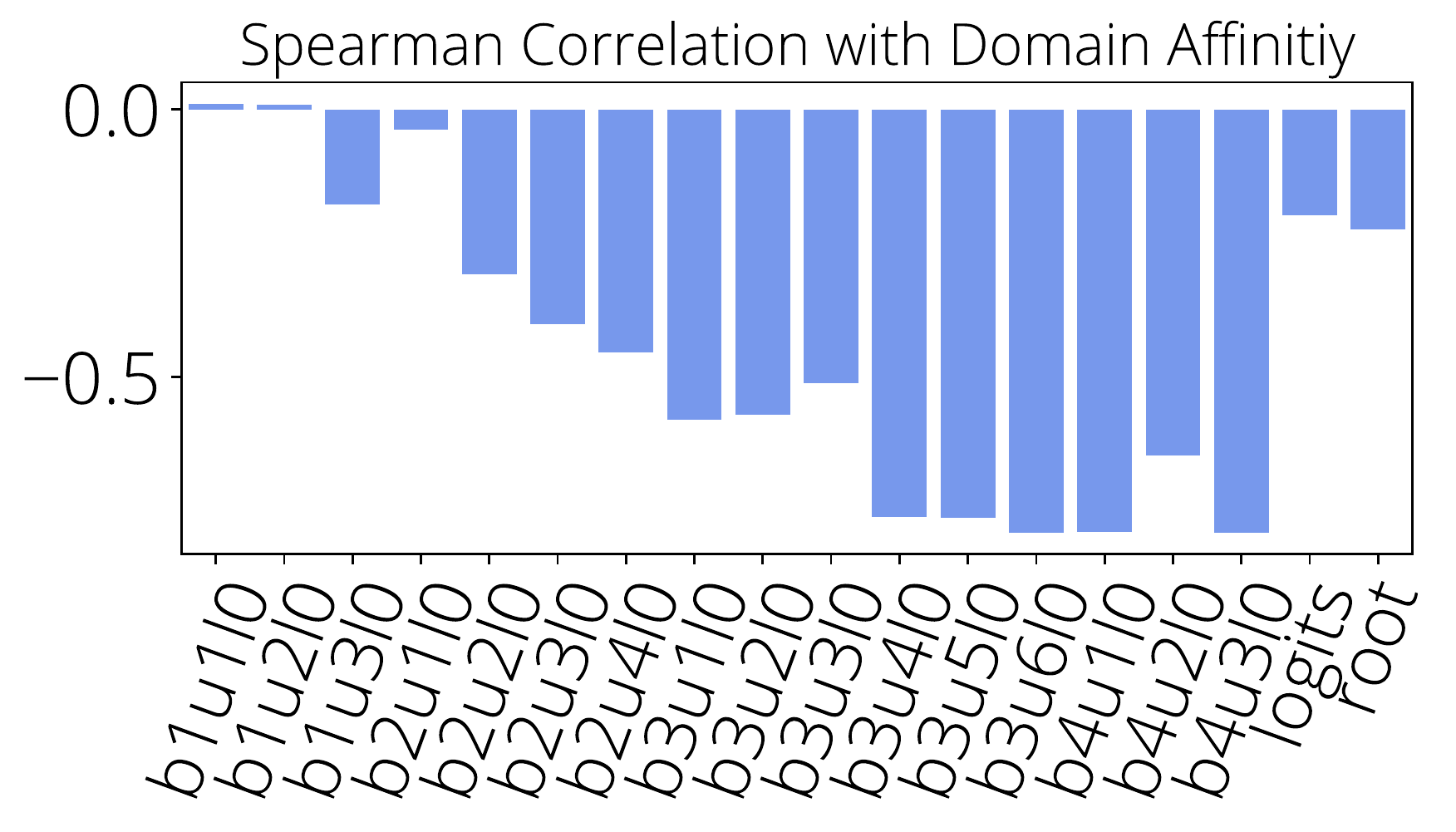}
\end{minipage}%
\\
\caption{\textbf{(left)} Domain affinity scores calculated using a ResNet-50 and a VIT-B/16 show a clear correlation. \textbf{(right)} For each layer, we calculate the the percentage improvement over LINEAR for each task. We report Spearman’s rank correlation between the percentage improvement and our domain affinity metric. All layers have a near-zero or negative correlation: the benefit from incorporating intermediate features diminishes with increased domain affinity, especially for the later layers.}
\label{app:fig:singlefeat}
\end{figure*}

\begin{figure*}[h]
\centering
\includegraphics[width=\columnwidth]{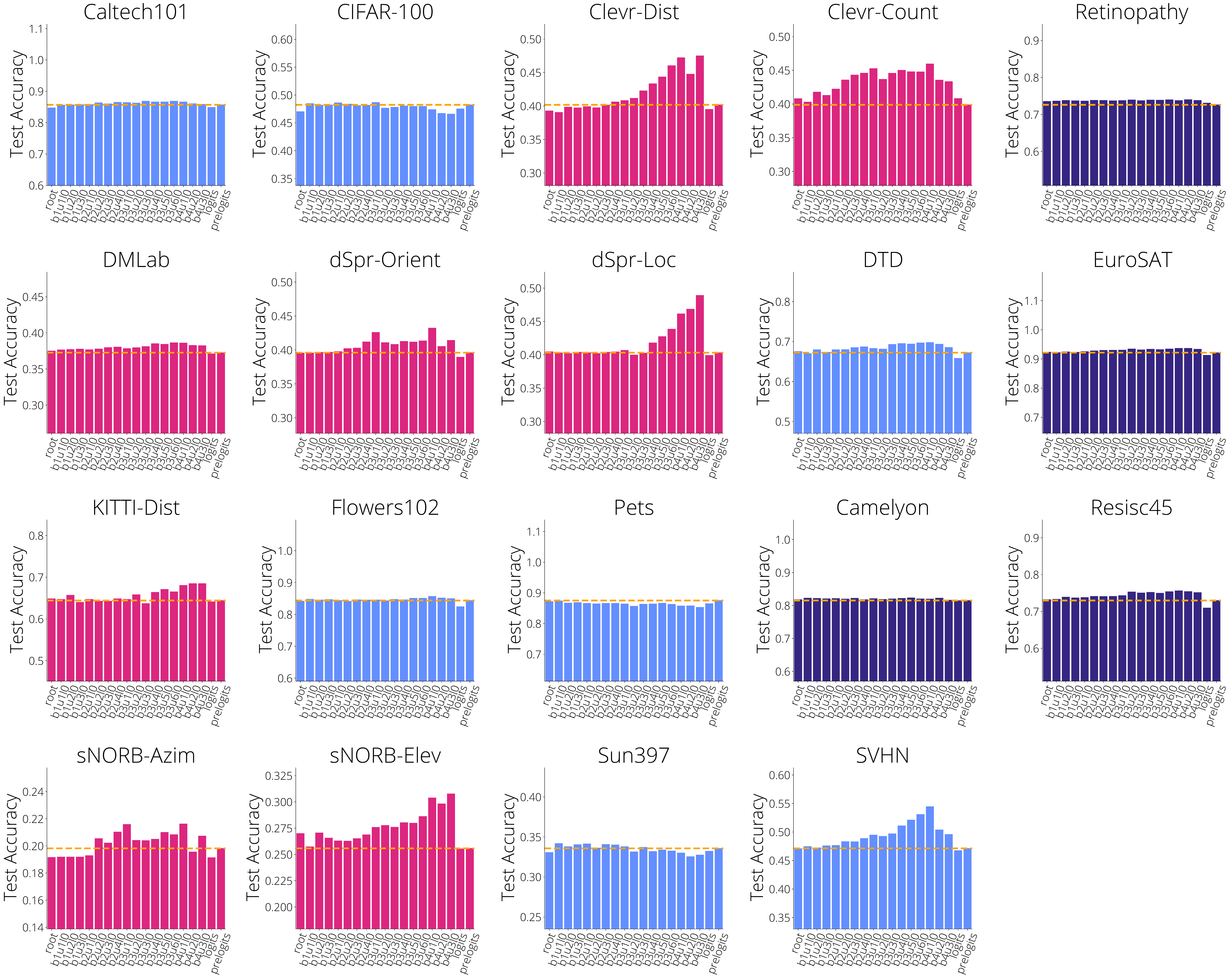}
\caption{Test accuracies when using a single additional intermedieate layer from a pretrained ResNet-50 backbone.}
\label{fig:app_oracle_full}
\end{figure*}

Test accuracies when using a single additional intermedieate layer from a pretrained ResNet-50
backbone are shown in \cref{fig:app_oracle_full}. Natural datasets (except SVHN) are highly similar to upstream dataset (ImageNet-2012) and thus adding an extra intermediate layer doesn’t improve performance much. However performance on OOD tasks (mostly of the tasks in structured category) improves significantly when intermediate representations are used, which correlates strongly with datasets in which \gls{h2t} exceeds \gls{ft} performance. This is also demonstrated in \cref{app:fig:singlefeat}-right, where we observe a negative correlation between domain similarity and accuracy gains for most of the intermediate layers when included during transfer. 

\newpage
\section{Additional Plots for Layer/Feature-wise Selection Comparison}
\label{app:fraction_full}
\begin{figure*}[h]
\centering
\includegraphics[width=\columnwidth]{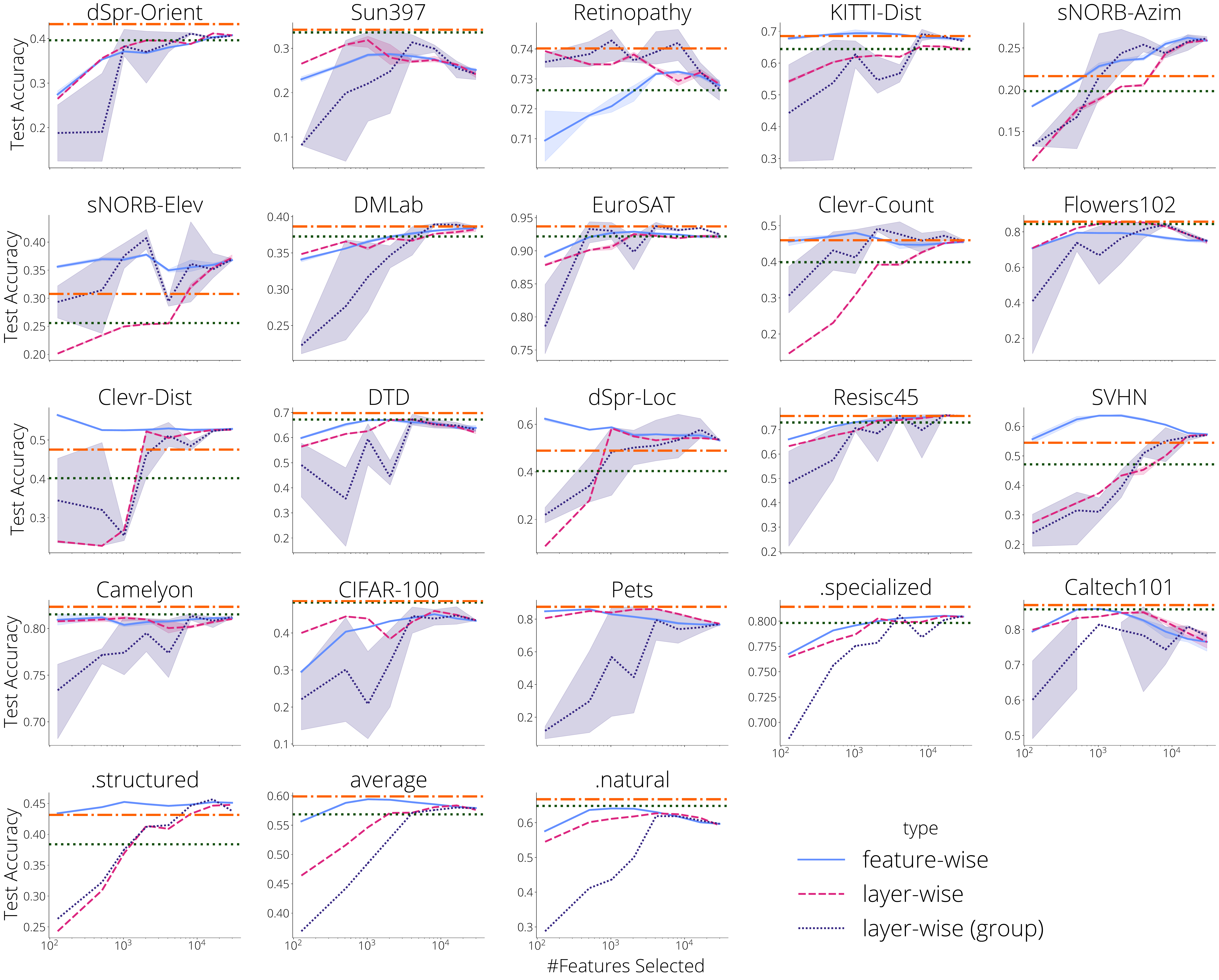}
\caption{Test accuracies when varying the number of features selected for \gls{h2t} using a pretrained ResNet-50 backbone.}
\label{fig:app_fraction_full}
\end{figure*}

We compare layer-wise selection strategy discussed in \cref{sec:exps:verify} to \gls{h2t} in \cref{fig:app_fraction_full}. For allmost all datasets, feature-wise selection produces better results. Retinopathy and Flowers-102 are the only two datasets where the layer-wise strategy performs better.

\begin{figure*}[h]
\centering
\begin{minipage}{.4\textwidth}
\centering
\includegraphics[width=\columnwidth]{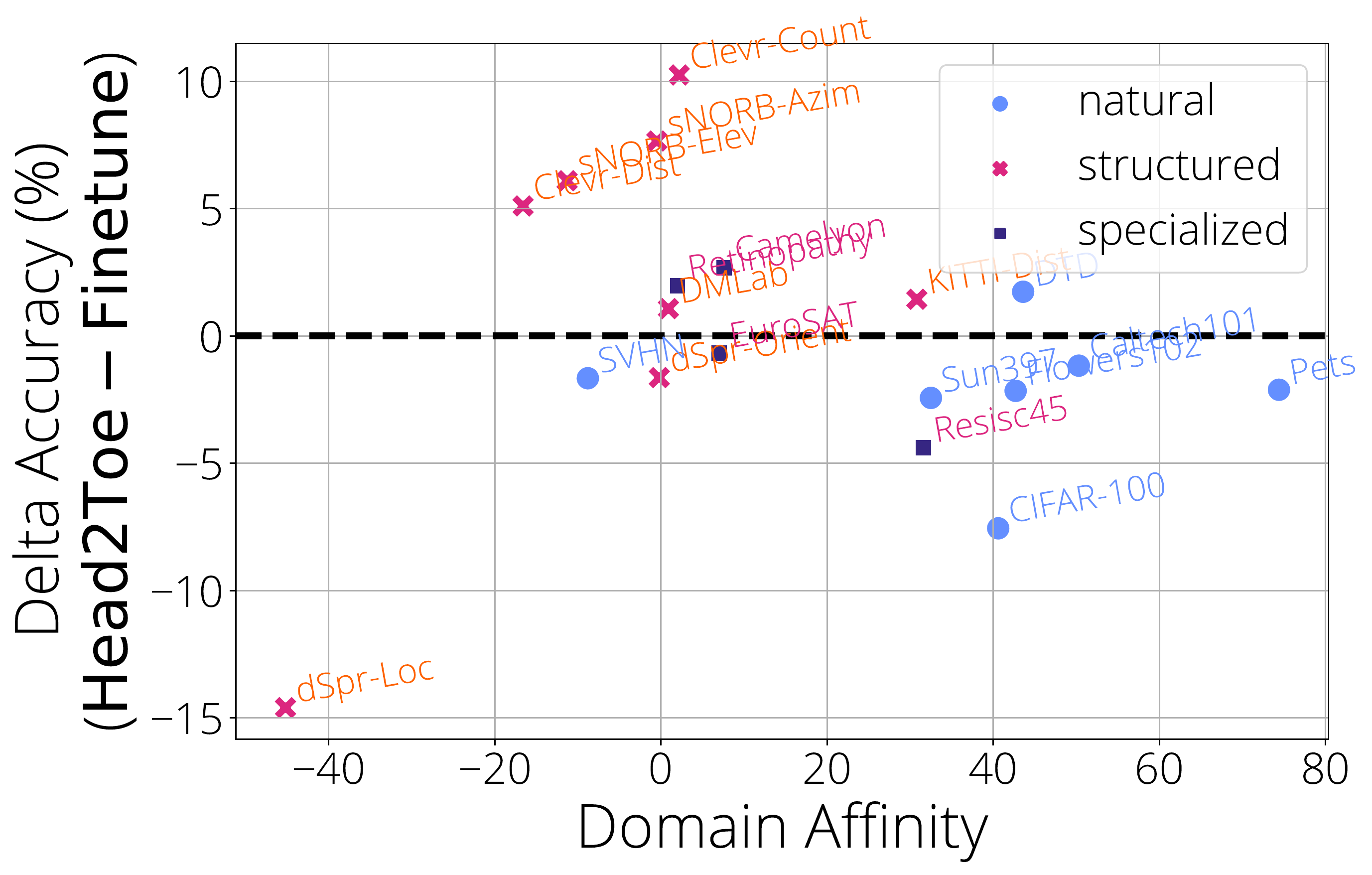}
\end{minipage}%
\begin{minipage}{.3\textwidth}
\centering
\includegraphics[width=\columnwidth]{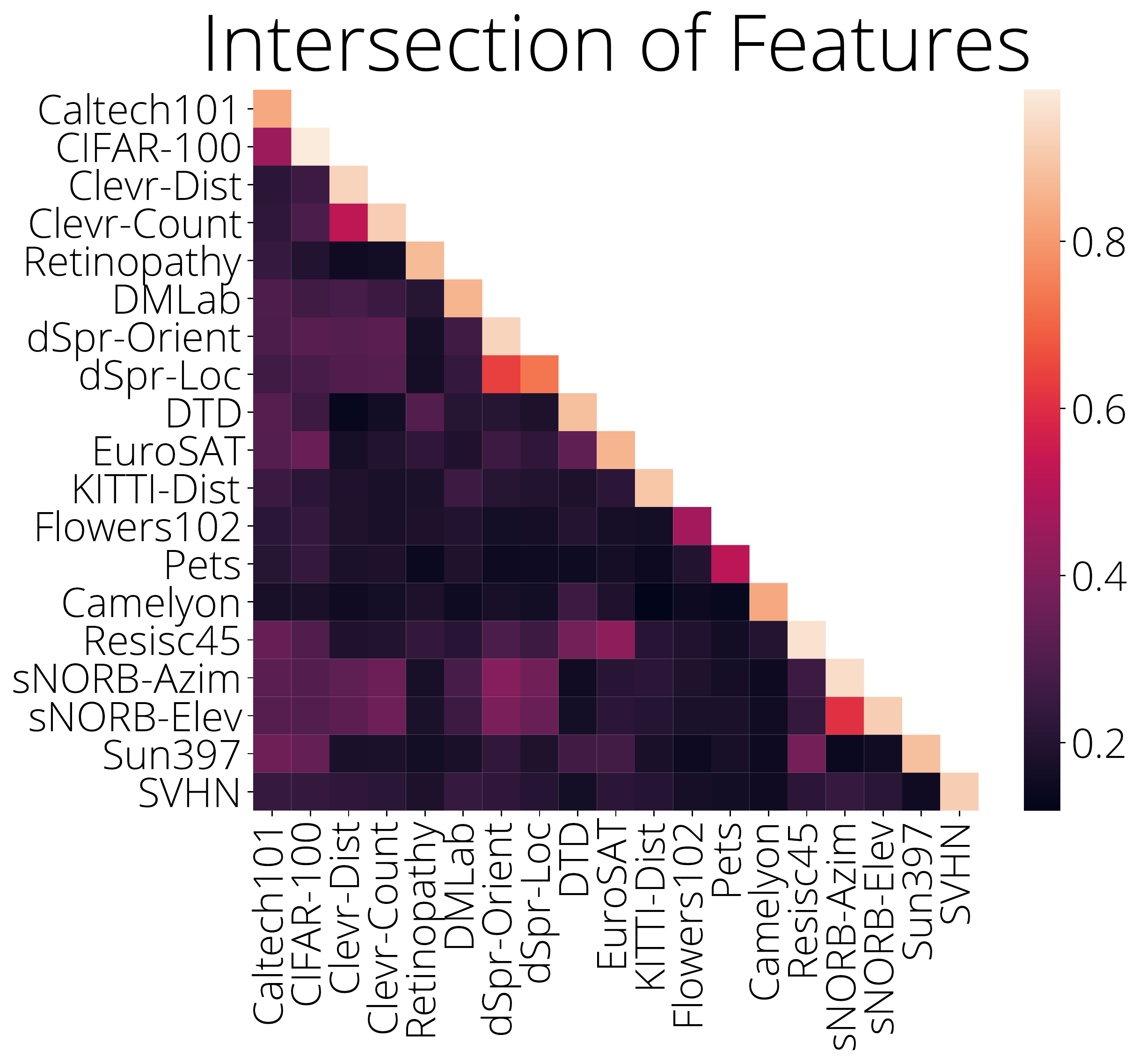}
\end{minipage}%
\begin{minipage}{.3\textwidth}
\centering
\includegraphics[width=\columnwidth]{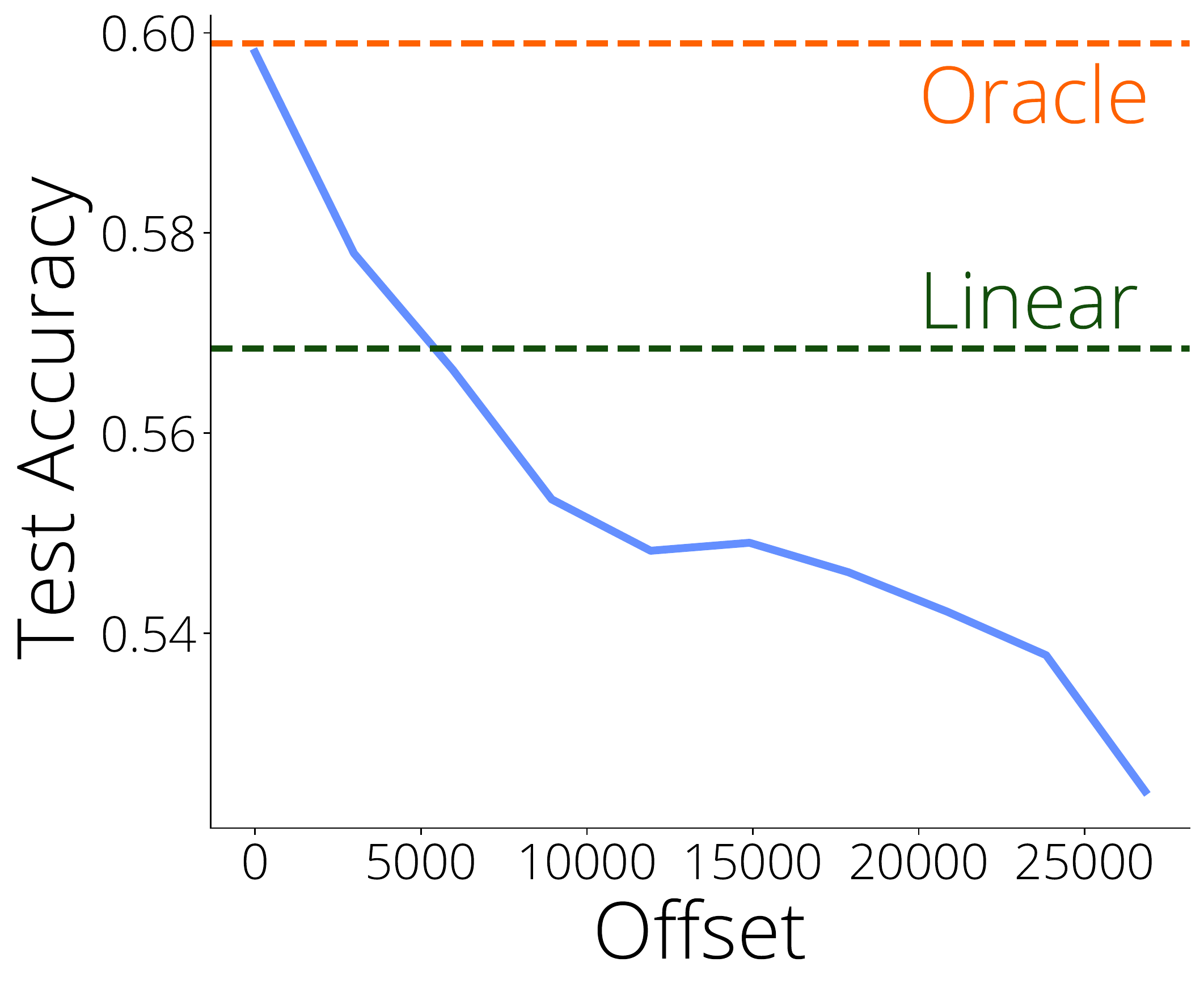}
\end{minipage}%
\caption{\textbf{(left)} Accuracy improvement of \gls{h2t} compared to \ft{}. \textbf{(center)} Intersection of features when selecting 2048 features from 29800 (same settings as in \cref{fig:h2t_demo}). The intersection is calculated as the fraction of features selected in two different runs. Values are averaged over 9 pairs of runs (3 seeds for each datasets), except the diagonal terms for which we remove identical pairs resulting in 6 pairs. \textbf{(right)} Over all VTAB tasks, average accuracy of \gls{h2t} when selecting 2048 consecutive features sorted by their relevance score, starting with an index specified by $\textit{offset}$.}
\label{fig:app:resnet_results}
\end{figure*}

\begin{figure*}[!htb]
\centering
\includegraphics[width=0.81\columnwidth]{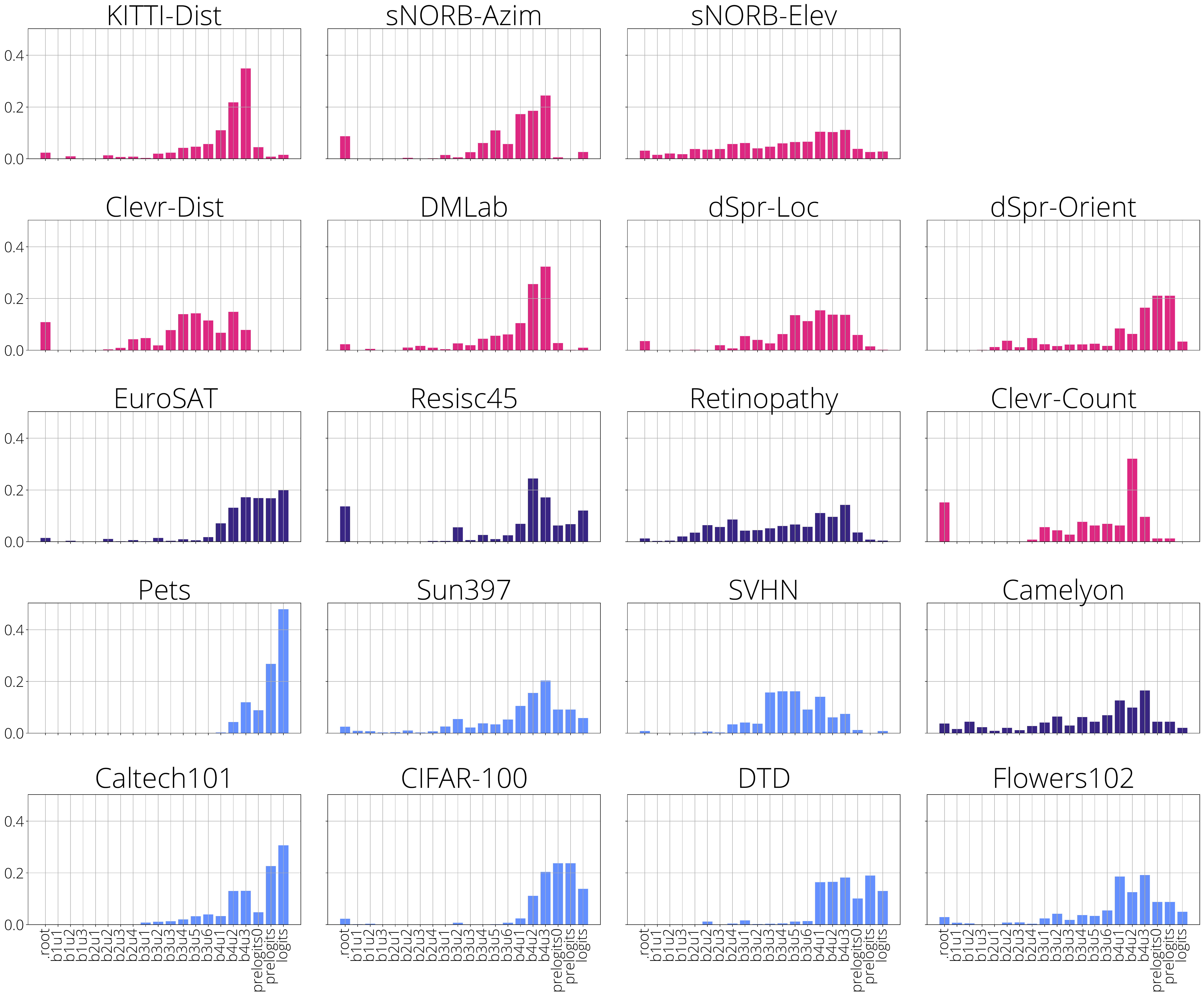}
\caption{Distribution of selected features over different ResNet-50 layers for VTAB-1k tasks for results presented in \cref{table:main}. We group the layers in each block (group of 3 layers) to reduce numbers of bars.}
\label{fig:resnet_dist_all}
\end{figure*}

\section{Additional Results for ResNet-50}
\label{app:resnet_results}
Improvement of \gls{h2t} over fine-tuning test accuracy for ResNet-50 backbone is shown in \cref{fig:app:resnet_results}-left. Similar to earlier plots, we observe a clear trend between being OOD and improvement in accuracy: \gls{h2t} obtains superior few-shot generalization for most of OOD tasks. We also share the distribution of features selected for each task in \cref{fig:resnet_dist_all}. Since different tasks might have different number features selected, we normalize each plot such that bars add up to 1. Overall, features from later layers seem to be preferred more often. Early layers are preferred, especially for OOD tasks like Clevr and sNORB. We observe a diverse set of distributions for selected features, which reflects the importance of selecting features from multiple layers. Even when distributions match, \gls{h2t} can select different features from the same layer for two different tasks. Therefore, next, we compare the indices of selected features directly to measure the diversity of features selected across tasks and seeds.   

\paragraph{Similarity of features selected.} In \cref{fig:app:resnet_results}-center we investigate the intersection of features selected by \gls{h2t}. We select 2048 features for each task from the pool of 29800 features (same experiments as in \cref{fig:h2t_demo}). For each task \gls{h2t} selects a different subset of features. We calculate the fraction of features shared between each subset. For each target task we run 3 seeds resulting in 3 sets of features kept. When comparing the similarity across seeds (diagonal terms), we average over 6 possible combinations among these 3 sets. Similarly, when comparing two different tasks we average over 9 possible combinations. Both dSprites and sNORB images are generated artificially and have similar characteristics. Results show that such complementary tasks have the highest overlap, around 40\%. Apart from a small fraction of tasks, however, most tasks seem to share less than 20\% of the features, which highlights, again, the importance of doing the feature selection for each target task separately.

\begin{figure}[h]
\centering
\includegraphics[width=\columnwidth]{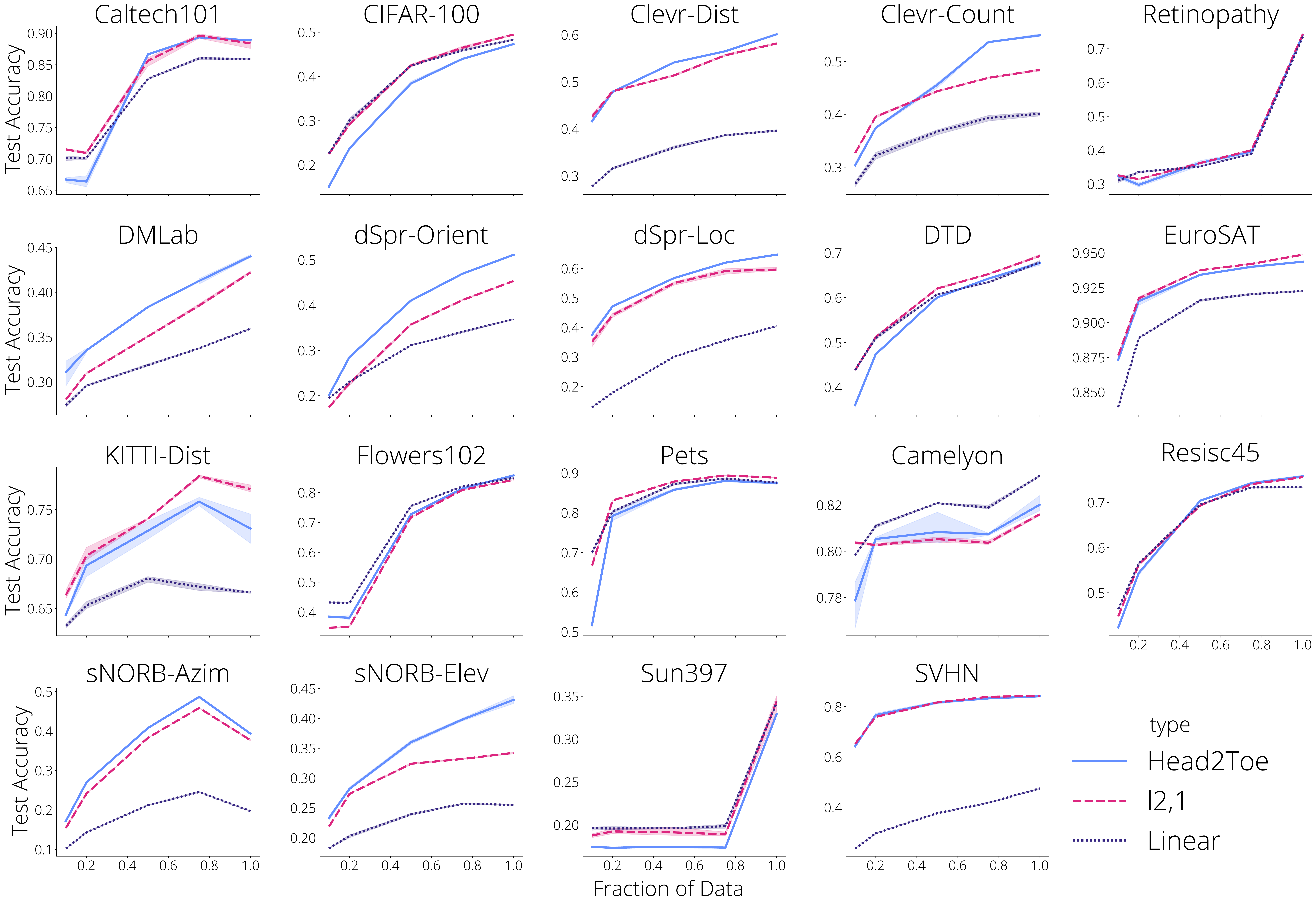}
\caption{Effect of data available during training to the test accuracy. Fraction=1 indicates original tasks with 1000 training samples. Overall we observe the performance of \gls{h2t} improves with amount of data available, possibly due to the reduced noise in feature selection.}
\label{fig:app_data_fractions_full}
\end{figure}

\paragraph{Effect of Training Data} In \cref{fig:app_data_fractions_full}, we compare the performance of \gls{h2t} with other baselines using reduced training data. Fraction ($f_d$)=1 indicates original tasks with 1000 training samples. For other fractions we calculate number of shots for each class as $int(1000*f_d/C)$ where $C$ is the number of classes and then sample a new version of the task accordingly. For SUN-397 task, number of shots are capped at 1 and thus fractions below 0.75 lead to 1-shot tasks and thus results are all the same. Overall we observe the performance of \gls{h2t} improves with amount of data available, possibly due to the reduced noise in feature selection.

\paragraph{Pre-activations or Activations?} Taylor approximation of the fine-tuning presented in \cref{subsec:taylor} suggests that the activations at every layer should be used. As an ablation we compare this approach with other alternatives in \cref{table:ablations:resnet}. Using only the CLS tokens at every layer without pooling (\textit{Only CLS}) or appending the CLS token to the pooled representations (\textit{Original+CLS}) didn't improve the results.

\paragraph{Including pooled input as a candidate for feature selection} We also try providing (possibly pooled) input to \gls{h2t}. As shown in \cref{table:ablations:resnet}, this didn't improved the results and we got around 0.8\% lower accuracy on average.

\paragraph{Relevance Scores.} In \cref{fig:app:resnet_results}-right, we demonstrate the effectiveness of group lasso on identifying relevant intermediate features of a ResNet-50 trained on ImageNet. We rank all features by their relevance score, $s_i$, and select groups of 2048 consecutive features beginning at a particular offset in this ranking. Offset 0 corresponds to selecting the features with largest relevance. We calculate average test accuracies across all VTAB tasks. As the figure shows, test accuracy decreases monotonically with the offset, indicating that the relevance score predicts the importance of including a feature in the linear classifier.

\section{Additional Results for ViT-B/16}
\label{app:vit_results}
\paragraph{Handling of CLS token} Class (CLS) tokens in vision transformers are often are added to the input and the classification layer is trained on top of the final representation of this token. Given that the representation of each token changes slowly, one might expect therefore the CLS representations along different layer to have more discriminate features. \gls{h2t} treats all tokens same and pools them together and in \cref{table:ablations:vit} we compare our approach with few other alternative approaches. Using only the CLS token at every layer without pooling (\textit{Only CLS}) or appending the CLS token to the pooled representations (\textit{Original+CLS}) doesn't improve the results.

\begin{figure*}[t]
\centering
\begin{minipage}{.35\textwidth}
\includegraphics[width=\columnwidth]{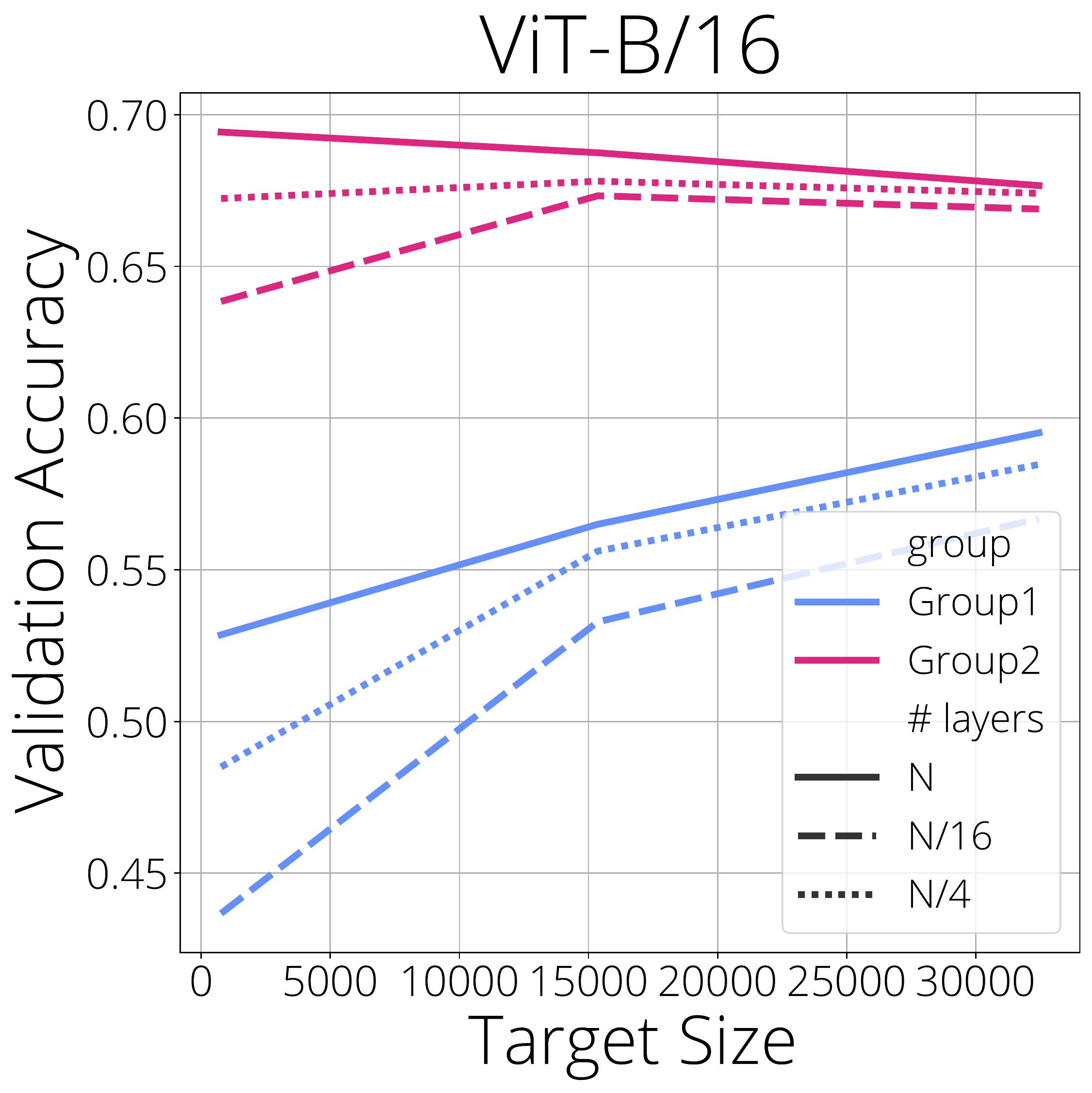}
\end{minipage}%
\begin{minipage}{.25\textwidth}
\fontsize{7pt}{7pt}\selectfont
\setlength{\tabcolsep}{1pt}
\setlength{\extrarowheight}{1pt}
\begin{tabular}{c|c}
Group1 & Group2 \\
\hline
DMLab& CIFAR-100 \\
DTD& Clevr-Count \\
sNORB-Azim& dSpr-Orient \\
SVHN& Retinopathy \\
dSpr-Loc& Resisc45 \\
Pets& EuroSAT \\
sNORB-Elev& Flowers102 \\
& Camelyon \\
& Caltech101 \\
& Clevr-Dist \\
& KITTI-Dist \\
\end{tabular}
\end{minipage}%
\caption{Effect of increasing the number of intermediate features that \gls{h2t} uses from the ViT-B/16 backbone. The abscissa of the graph indicates the dimensionality of the representation extracted from each layer of the backbone (\emph{target size}). The tasks are split into two groups (see right side of Figure), which show different behavior. The solid, dashed, and dotted lines indicate the fraction of layers selected for forming the representation used by \gls{h2t}: 1/16, 1/4, and 1, respectively. Scaling curves for individual tasks can be found in \cref{app:scaling_all}. Sun397 experiments with all layers and largest target feature size (24576) failed due to memory issues thus we don't include Sun397 results in aggregated plots}
\label{fig:app:vit_scale}
\end{figure*}

\paragraph{Scaling of ViT-B/16} We repeat the scaling plot for ResNet-50 (\cref{fig:understanding}-right) for ViT-B/16 at \cref{fig:app:vit_scale} and observe a quite similar scaling behaviour for the two groups of datasets.

\begin{table}[t]
\centering
\begin{tabular}{lllll}
\toprule
&  Average &  Natural &  Specialized & Structured \\
\midrule
Original         &  65.8 &     70.5 &         81.6 &        53.7 \\
\hline
Pre-Normalization &  63.5 &     67.4 &         81.1 &        51.3 \\
Pre-Activation  &  63.5 &     67.4 &         81.3 &        51.3 \\
\hline
Adding Pooled Input  &  65.0 &     69.7 &         81.2 &        52.9 \\
\bottomrule
\end{tabular}

\caption{Ablation of choice of activations used on \gls{h2t} results for ResNet-50 backbone. Using features before the activation (\textit{Pre-Activation}) or the normalization (\textit{Pre-Normalization}) or adding the inputs to the pool of features considered bring worse results.}
\label{table:ablations:resnet}
\end{table}

\begin{table}[!h]
\centering
\begin{tabular}{lrrrr}
\toprule
&  Average &  Natural &  Specialized & Structured \\
\midrule
Original                                           &  63.3 &     71.3 &         82.6 &        46.7 \\
\hline
Only CLS                                           &  57.3 &     65.1 &         81.4 &        38.4 \\
Original+CLS                                       &  63.4 &     72.0 &         82.0 &        46.6 \\
\hline

\bottomrule
\end{tabular}
\caption{Alternatives treatments for the CLS token of the ViT-B/16 architecture. Using only the CLS tokens at every layer without pooling (\textit{Only CLS}) or appending the CLS token to the pooled representations (\textit{Original+CLS}) doesn't improve the results.}
\label{table:ablations:vit}
\end{table}
\newpage

\begin{figure}[h]
\centering
\includegraphics[width=0.9\columnwidth]{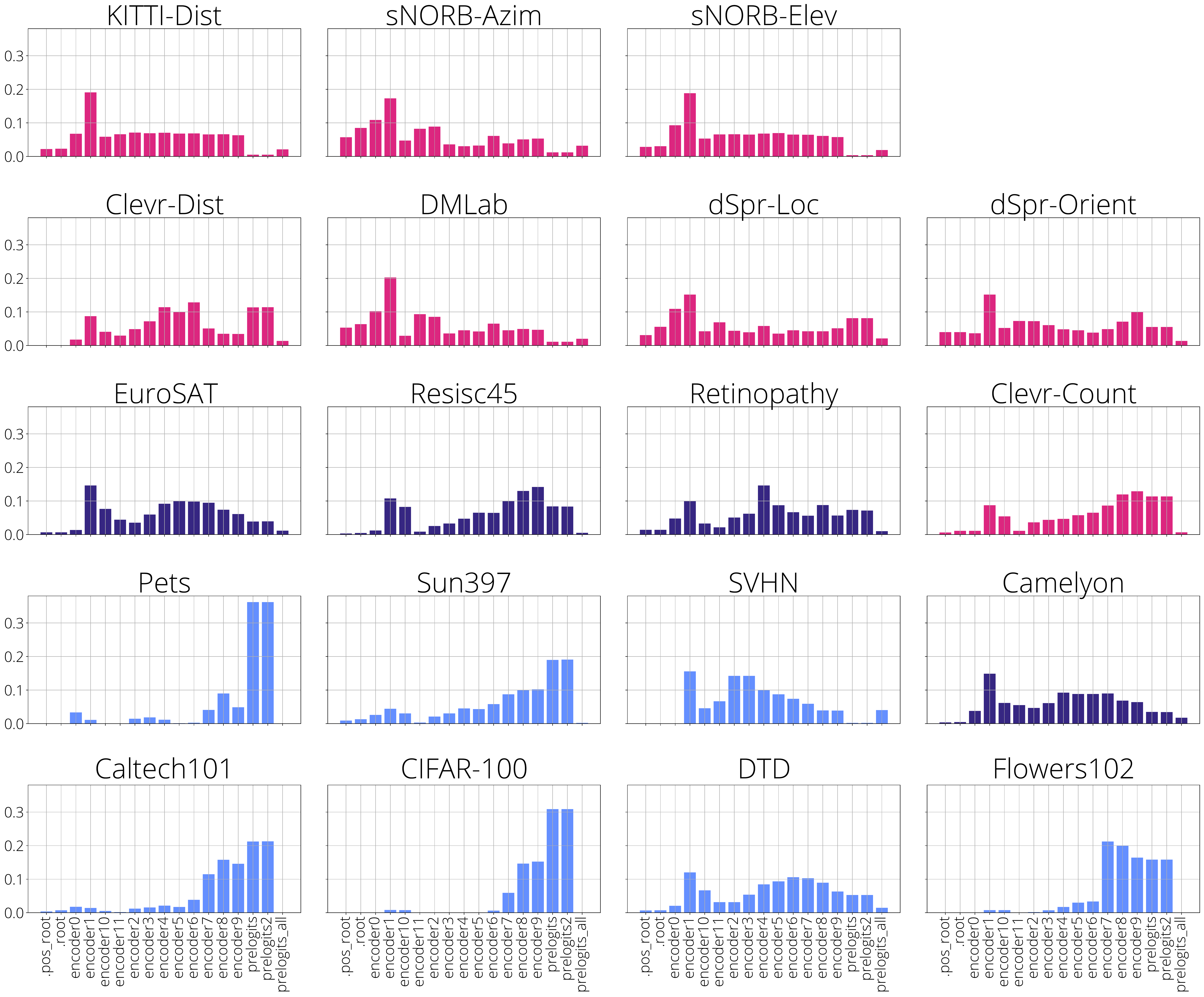}
\\
\caption{Distribution of selected features over different ViT-B/16 layers for VTAB-1k tasks for results presented in \cref{table:main}. We group the layers in encoder (group of 4 activations) to reduce numbers of bars.}
\label{fig:app:vit_dist_all}
\end{figure}

\paragraph{Distribution of features across layers} We share the distribution of features selected for each task in \cref{fig:app:vit_dist_all}. As before, we normalize each plot such that bars add up to 1. Similar to ResNet-50, features from later layers selected more often for tasks from natural category. Early layers are preferred, especially for OOD tasks. In general distributions are more balanced, in other words, more intermediate features are used compared to ResNet-50.

\newpage
\section{Additional Plots for Scaling Behaviour of Head2Toe}
\label{app:scaling_all}
\begin{figure}[!th]
\centering
\includegraphics[width=0.78\columnwidth]{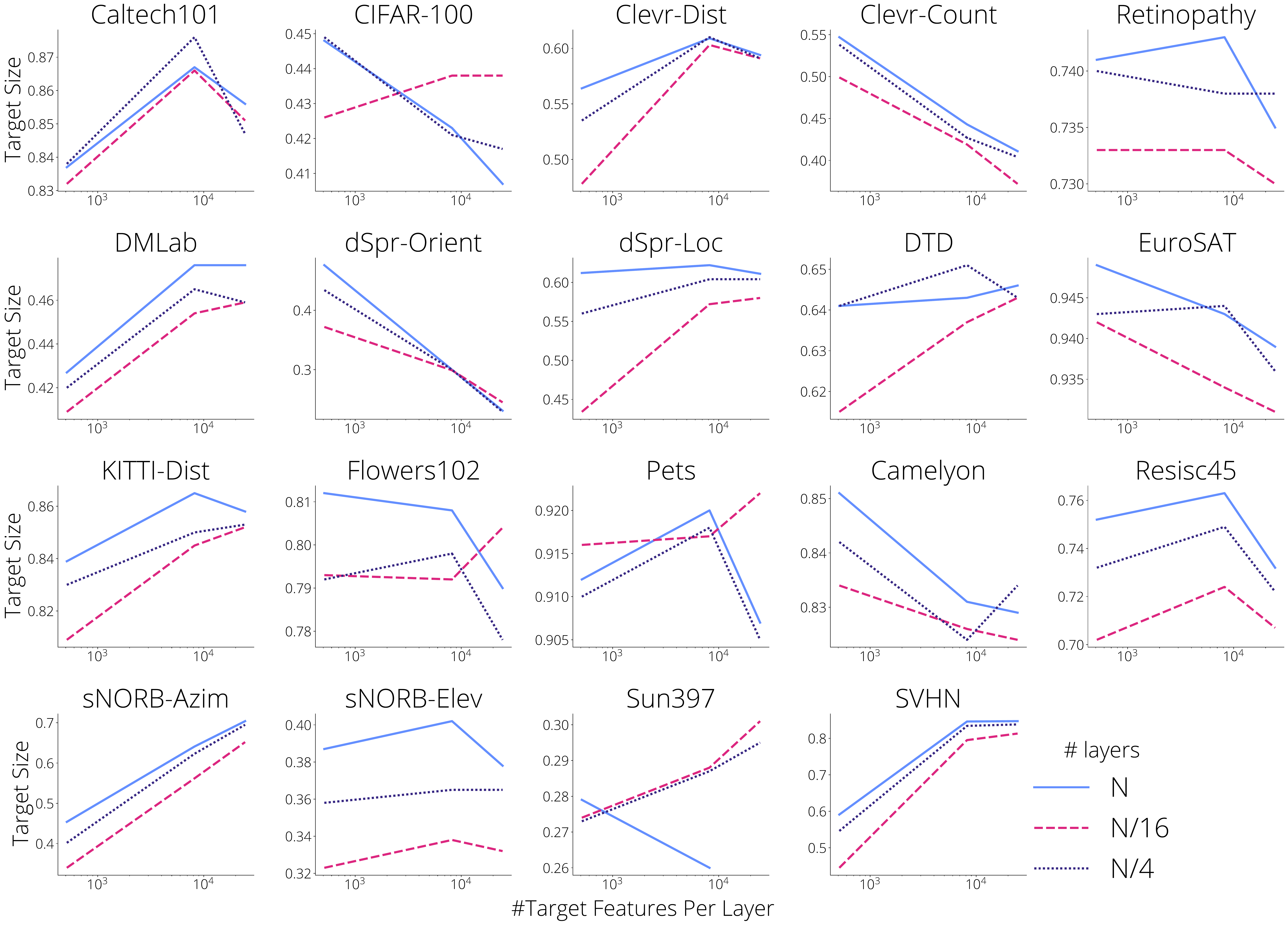}
\includegraphics[width=0.78\columnwidth]{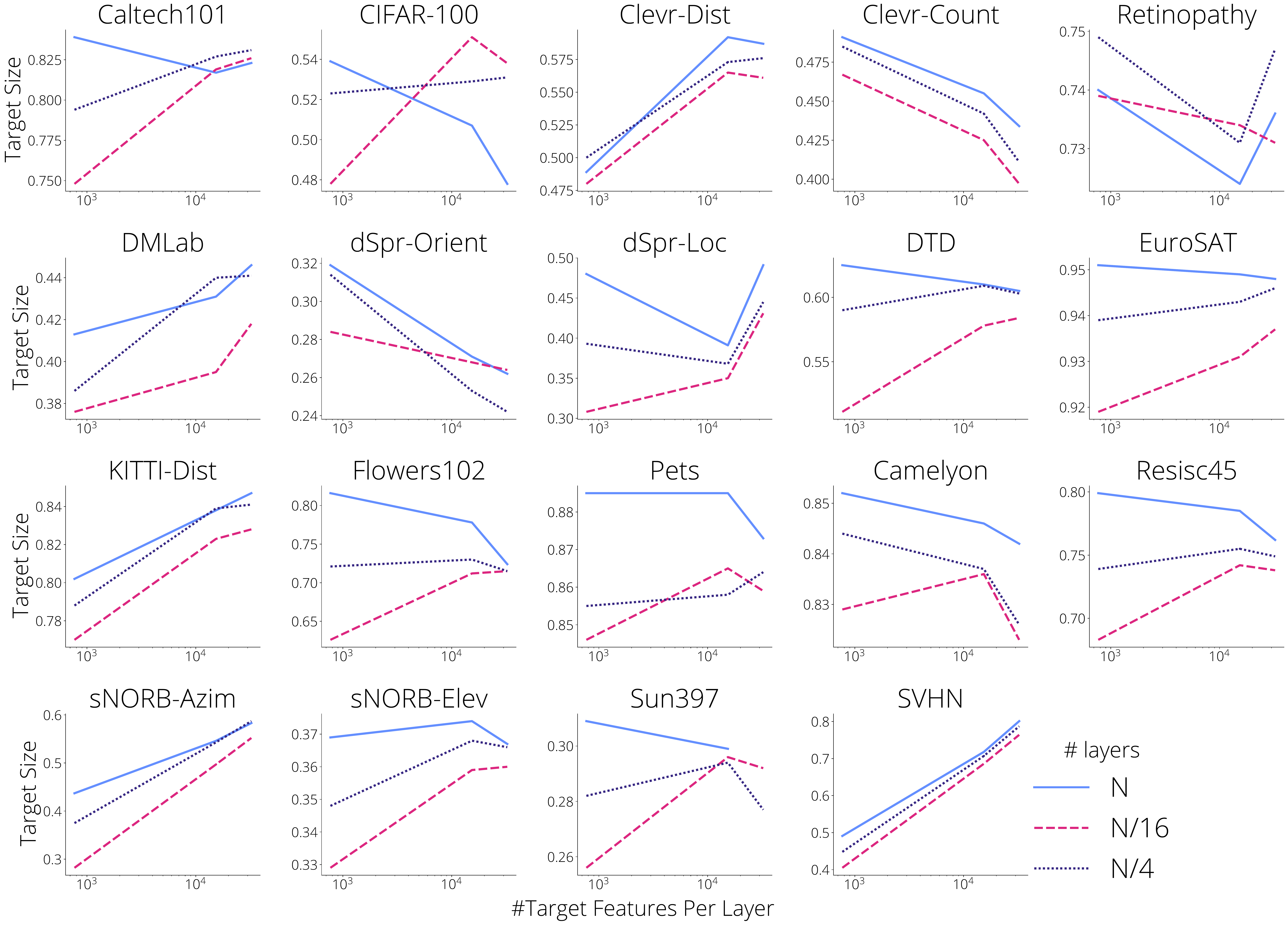}
\caption{Scaling Behaviour over 19 VTAB-1k tasks when varying feature target size and number of layers utilized for \textbf{(top)} ResNet-50 and \textbf{(bottom)} ViT-B/16. Sun397 experiments with all layers and largest target feature size (24576) failed due to memory issues.}
\label{fig:app_scaling_all}
\end{figure}

Scaling behaviour of \gls{h2t} when using different feature target size and number of layers over 19 VTAB-1k tasks is shown in \cref{fig:app_scaling_all}. Sun397 experiments with all layers and largest target feature size (24576) failed due to memory issues thus we don't include Sun397 results in aggregated plots.


\end{document}